\documentclass[acmsmall,screen]{acmart}

\setcopyright{acmlicensed}
\copyrightyear{2025}
\acmYear{2025}
\acmDOI{XXXXXXX.XXXXXXX}
\acmISBN{978-1-4503-XXXX-X/xxxx/xx}

\usepackage{amsmath,amsfonts}
\usepackage{pifont}

\usepackage{soul}

\usepackage{algpseudocode}
\usepackage{algorithm}

\usepackage{graphicx}
\graphicspath{ {figures/} }
\usepackage[caption=false,font=normalsize,labelfont=sf,textfont=sf]{subfig}

\usepackage{tikz}
\usepackage{pgfplots}
\pgfplotsset{compat=1.18}
\usepackage{pgfplotstable}
\usepackage{stackengine}   

\usepackage{tcolorbox}
\usepackage{adjustbox}

\usepackage{tabularx}
\usepackage{booktabs}
\usepackage{threeparttable} 
\usepackage{multirow}
\usepackage{array}

\usepackage{cprotect}

\usepackage[table]{xcolor}
\definecolor{lightred}{RGB}{255, 128, 128}
\definecolor{lightorange}{RGB}{255, 192, 128}
\definecolor{lightyellow}{RGB}{255, 255, 128}
\definecolor{lightgreen}{RGB}{128, 255, 128}
\definecolor{lightblue}{RGB}{128, 192, 255}
\definecolor{lightpurple}{RGB}{230, 230, 250}
\definecolor{lightpink}{RGB}{255, 192, 203}

\usepackage{xurl}

\usepackage{enumitem}
\usepackage{float}
\usepackage{listings}
\newfloat{lstfloat}{htbp}{lop}
\floatname{lstfloat}{Listing}
\lstdefinestyle{xmlstyle}{
    language=XML,
    basicstyle=\ttfamily\footnotesize,
    morestring=[b]`,
    morecomment=[s]{<!--}{-->},
    commentstyle=\color{gray}\upshape,
    stringstyle=\color{orange},
    identifierstyle=\color{black},
    keywordstyle=\color{cyan},
    breaklines=true,
    breakatwhitespace=true,
    tabsize=2
}
\lstdefinelanguage{json}{
    basicstyle=\ttfamily\footnotesize,
    showstringspaces=false,
    breaklines=true,
    frame=lines,
    backgroundcolor=\color{gray!10},
    morestring=[b]`,
    stringstyle=\color{brown},
    literate=
     *{0}{{{\color{blue}0}}}{1}
      {1}{{{\color{blue}1}}}{1}
      {2}{{{\color{blue}2}}}{1}
      {3}{{{\color{blue}3}}}{1}
      {4}{{{\color{blue}4}}}{1}
      {5}{{{\color{blue}5}}}{1}
      {6}{{{\color{blue}6}}}{1}
      {7}{{{\color{blue}7}}}{1}
      {8}{{{\color{blue}8}}}{1}
      {9}{{{\color{blue}9}}}{1}
}

\definecolor{capitalblack}{HTML}{000000}
\newcommand{\CAP}[1]{\textbf{\textcolor{capitalblack}{#1}}}

\newcommand{\RomanNumber}[1]{\uppercase\expandafter{\romannumeral #1\relax}}

\newcommand\SysName{LADFA}

\begin{document}

\title[\SysName: A Framework of Using LLMs and RAG for Personal Data Flow Analysis in Privacy Policies]{\SysName: A Framework of Using Large Language Models and Retrieval-Augmented Generation for Personal Data Flow Analysis in Privacy Policies}

\author{Haiyue Yuan}
\email{h.yuan-221@kent.ac.uk}
\orcid{0000-0001-6084-6719}
\authornote{Corresponding co-authors.}
\affiliation{%
  \institution{Institute of Cyber Security for Society (iCSS) \& School of Computing, University of Kent}
  \city{Canterbury}
  \country{United Kingdom}
}

\author{Nikolay Matyunin}
\email{nikolay.matyunin@honda-ri.de}
\orcid{0000-0001-8974-3078}
\author{Ali Raza}
\email{ali.raza@honda-ri.de}
\orcid{0000-0001-8326-8325}
\affiliation{%
  \institution{Honda Research Institute Europe GmbH}
  \country{Germany}
}

\author{Shujun Li}
\authornotemark[1]
\email{s.j.li@kent.ac.uk}
\orcid{0000-0001-5628-7328}
\affiliation{%
  \institution{Institute of Cyber Security for Society (iCSS) \& School of Computing, University of Kent}
  \city{Canterbury}
  \country{United Kingdom}
}

\renewcommand{\shortauthors}{Yuan et al.}

\begin{abstract}
Privacy policies help inform people about organisations' personal data processing practices, covering different aspects such as data collection, data storage, and sharing of personal data with third parties. Privacy policies are often difficult for people to fully comprehend due to the lengthy and complex legal language used and inconsistent practices across different sectors and organisations. To help conduct automated and large-scale analyses of privacy policies, many researchers have studied applications of machine learning and natural language processing techniques, including large language models (LLMs). While a limited number of prior studies utilised LLMs for extracting personal data flows from privacy policies, our approach builds on this line of work by combining LLMs with retrieval-augmented generation (RAG) and a customised knowledge base derived from existing studies. This paper presents the development of \SysName, an end-to-end computational framework, which can process unstructured text in a given privacy policy, extract personal data flows and construct a personal data flow graph, and conduct analysis of the data flow graph to facilitate insight discovery. The framework consists of a pre-processor, an LLM-based processor, and a data flow post-processor. We demonstrated and validated the effectiveness and accuracy of the proposed approach by conducting a case study that involved examining ten selected privacy policies from the automotive industry. Moreover, it is worth noting that \SysName\ is designed to be flexible and customisable, making it suitable for a range of text-based analysis tasks beyond privacy policy analysis.
\end{abstract}




\maketitle

\section{Introduction}
\label{sec:introduction}

Privacy policies are widely used as a means of informing people about personal processing data practices of organisations. They are considered essential documents that detail information about how organisations collect, share, and manage personal data of people (i.e., data subjects in legal terms). Privacy policies are often subject to legal requirements, such as the General Data Protection Regulation (GDPR) in the EU~\cite{EU-GDPR} and the UK~\cite{UK-GDPR} and various state-level regulations in the US (e.g., the California Consumer Privacy Act (CCPA)~\cite{CCPA} and the California Privacy Rights Act (CPRA)~\cite{CALOPPA}). While these regulations mandate organisations to disclose specific information, their implementation often lacks consistency and transparency~\cite{Schaub-F2016, Karl-2024}. Moreover, due to the lengthy and sophisticated legal language used in privacy policies, human readers often struggle to fully understand the scale and scope of data collection and sharing practices or often choose not to read them at all~\cite{McDonald-A2008, Cate-F2010, Tang-P2024}.

To address these challenges, some researchers have examined privacy policy consistency, clarity, and transparency to better inform and present meaningful information to human readers. Machine learning (ML) and natural language processing (NLP) techniques have been frequently explored to facilitate such analyses~\cite{Fabian-B2017, Oltramari-A2018, Harkous-H2018, Andow-B2019}. More recently, with the increasing capabilities and popularity of large language models (LLMs), several studies have shown promising results of leveraging LLMs to automate the privacy policy analysis process in zero- and/or few-shot training contexts~\cite{Tang-C2023, Salvi-RC2024, Goknil-A2024, Chen-Y2025, Mori-K2025}.

Among all information described in a privacy policy, personal data flows are of particular importance as they can tell data subjects what personal data about them will be collected by whom and with what third parties the collected personal data will be shared, under which conditions and for what purposes. Such personal data flows can be understood as information flows in the contextual integrity (CI) theory, proposed by \citet{Nissenbaum-H2009}. In the CI theory, privacy can be modelled as personal data flows, which are governed by contextual norms. The key parameters of such norms are \emph{actors} (including the \emph{subject} of information, the \emph{sender} of information, and \emph{recipient} of information), \emph{attributes} (i.e., types of information), and \emph{transmission principle}~\cite{Nissenbaum-H2009, Nissenbaum-H2011}, which refers to the condition or constraints that govern the personal data flow, restricting it to specific circumstances (e.g., a business should only give customer records to the government if there is a warrant or court order)~\cite{Malkin-N2023}.

These parameters are essential to inform data subjects about the personal data collected, with whom it is shared, and for what purposes. Although `purposes' is not explicitly defined in the original CI theory, \citet{Nissenbaum-H2019} later explained that factors such as purpose do not exist in a context but help constitute it. In this sense, we consider purpose as part of \emph{transmission principles}. For the readability and consistency, we refer to personal data flow as data flow for the rest of this paper and refer to data flows that include these attributes as comprehensive data flows. Although a few studies have explored manual~\cite{Yuan-H2023, Ghahremani-S2024} and automatic approaches~\cite{Cui-H2023, Xie-Q2025, Yang-M2025} to extract data flows from privacy policies and construct data flow graphs for visualisation and insights discovery, to the best of our knowledge, no existing study has utilised LLMs with RAG and a customised knowledge base to automatically extract comprehensive data flows from the perspective of CI theory.

To address the above-mentioned research gap, we conducted a study with following main contributions: 
\begin{itemize}
    \item \textbf{Novel framework}: we propose an end-to-end privacy policy analysis framework \SysName\ (short for ``A Framework using \textbf{L}LMs and R\textbf{A}G for Personal \textbf{D}ata \textbf{F}low \textbf{A}nalysis in Privacy Policies''), focusing on extracting data flows, constructing and analysing data flow graphs. It consists of a pre-processor, an LLM-based processor, and a data flow post-processor. 
    \item \textbf{Customised knowledge base}: beyond utilising LLMs, we design and construct a knowledge base for providing useful contextual information that allow the framework to leverage retrieval-augmented generation (RAG) to facilitate the LLM-based processor and the post-processor. 
    \item \textbf{Case study with evaluation}: we conduct a case study on ten connected-vehicle mobile apps from different original equipment manufacturers (OEMs) in the automotive industry. Unlike prior work that evaluated only segments of data flows separately, we provide a collective evaluation of comprehensive data flows. Due to the lack of ground truth, manual validation was performed by three domain experts (the first three co-authors of the paper). The evaluation results showed strong agreement with the \SysName's outputs, with the average 7-Likert scores between 6 and 7 for most tasks. Gwet's AC1 and percentage agreement for identifying data types and data flows reached 0.94 and 0.82, and 0.96 and 0.86, respectively, indicating high inter-rater reliability. These results affirm \SysName's capability in processing, understanding unstructured texts, and extracting data flows from privacy policies.
    \item \textbf{Insights discovery}: the comparison of data flow graphs and graph network analysis demonstrated \SysName's capability in analysing and discovering privacy- and security-related insights that are often difficult to comprehend or easily overlooked by human readers of privacy policies.

\end{itemize}

The rest of this paper is organised as follows. Related work is discussed in Section~\ref{sec:related_work}. Section~\ref{sec:framework} provides detailed descriptions of each component of the proposed end-to-end framework for automated privacy policy analysis. Section~\ref{sec:case_study} presents the details of the case study and the results, followed by a discussion of some identified limitations in Section~\ref{sec:limitations}. Finally, the last section concludes the paper.

\section{Related Work} 
\label{sec:related_work}

\subsection{Privacy Policy Analysis in General}

Previous research has indicated that a significant proportion of organisations or services do not adequately inform users about data-handling practices through privacy policies. A large-scale analysis conducted in 2017 highlighted that approximately 50\% of popular free apps available on the Google Play app store were missing a privacy policy, despite the fact that a large portion of them (70\%) are capable of processing personally identifiable information~\cite{Leigh-S2017}. Similarly, another study~\cite{Robillard-JM2019} found that most smartphone apps do not include privacy policies: for the privacy policies that were included, only 18\% of the iOS apps could be accessed, and accessibility was even lower for the Android apps (approximately 4\%). Apart from such accessibility challenges, known issues related to the readability, presentation, transparency, and consistency of privacy policies often prevent more effective and informative communication to consumers~\cite{Cottrill-C2013, Robillard-JM2019, Chen-B2023}. Consequently, consumers rarely read privacy policies and perceive them as overly complex, lengthy, and difficult to comprehend, resulting in uninformed consent to data practices~\cite{Wilson-S2018, Linden-T2020, Oltramari-A2018, Yuan-H2023, Karl-2024}. Such challenges not only undermine user trust but also complicate regulatory compliance for organisations~\cite{Fabian-B2017}. More recently, \citet{Ghahremani-S2024} presented a study to systematically evaluate the transparency and comprehensiveness of privacy policies by manually annotating and identifying contextual gaps and ambiguities based on the CI framework. They argued that such manual analysis can be considered an alternative to subjective evaluations by privacy experts; however, there is a pressing need to automate the process using ML and NLP-based approaches, especially for large-scale studies and for developing an automated tool to assist human users.

\subsection{Automated Privacy Policy Analysis}

\subsubsection{ML and NLP-based Approaches}

In addition to the challenges listed above, it is essential to analyse privacy policy content to evaluate its completeness and alignment with regulatory frameworks and improve its readability, presentation, consistency, and clarity. Given the complexity and volume of text in privacy policies, researchers have explored various automated approaches that utilise ML and NLP techniques for such analyses.

Several studies have investigated and confirmed the readability issues of privacy policies from different domains. \citet{Fabian-B2017} developed an automated toolset that utilised NLP techniques for information extraction and readability analysis. They examined 50,000 privacy policies from popular English-speaking websites and adopted the Flesh-Kincaid Grade Level (FKG) metric, a well-known metric for evaluating readability. The mean FKG score was 13.6, indicating that, on average, these documents remain difficult for human readers to comprehend. In a different study, \citet{Srinath-K2021} presented the PrivaSeer corpus, which contains 1,005,380 privacy policies from 995,475 web domains. The readability analysis of this dataset yielded a mean FKG score of 14.87, suggesting a need for roughly two years of US college education to fully grasp a typical privacy policy. In another study on the privacy policies of mental health apps, \citet{Robillard-JM2019} calculated average scores from three established readability metrics, including Gunning Fog, FKG, and the Simple Measure of Gobbledygook (SMOG). The results reveal that these legal documents require a reading level equivalent to that of a college student, making them difficult for the average user to comprehend fully.

Such readability issues are often caused by complex legal language and the manner in which the content is presented. Numerous studies have focused on bridging this gap using various approaches, including innovative visual representations, nudging techniques, question-answering systems, and summarisation tools, to facilitate more effective and informative communication with consumers. For instance, \citet{Oltramari-A2018} proposed PrivOnto, a semantic framework that combines crowdsourcing, ML and NLP that can represent annotated privacy policies using an ontology, allowing the development of SPARQL queries to extract information from the PrivOnto knowledge base to address user privacy-related questions and assist researchers and regulators in large-scale privacy policy analysis. \citet{Zaeem-RN2018} developed a Chrome browser extension, PrivacyCheck, which utilises trained data mining classification models using 400 privacy policies. It allows summarising an HTML-based privacy policy and presents the results as graphical icons with short descriptions and risk-level indications. Similarly, \citet{Harkous-H2018} developed an automated framework, Polisis, which leverages a privacy-centric language model trained on 130 K privacy policies and a neural network classifier to analyse both high-level and fine-grained details of privacy practices. Polisis can automatically assign privacy icons to privacy policies, allowing human readers to learn more privacy insights for making more informed decisions. In addition, \citet{Harkous-H2018} introduced PriBot based on Polisis. It is the first interactive question-answering system for privacy policies, offering consumers a more dynamic and engaging way to understand these documents. Furthermore, \citet{Kumar-B2020} conducted a study leveraging ML algorithms to extract clean texts specifically related to opt-out options. This work further facilitates the development of a browser extension designed to help people better understand their opt-out choices. More recently, \citet{Bui-D2021} presented an automated system, PI-Extract, which uses a neural network model to extract privacy practices. In the same paper, a follow-up user study on investigating the effects of data practice annotations to highlight the extracted privacy practices using PI-Extract was reported to help human readers better comprehend privacy policies.

In addition to the research directions mentioned above on automated privacy policy content analysis, various studies have focused on exploring the inconsistency, lack of clarity, and misalignment of privacy policy content with the regulatory frameworks. \citet{Andow-B2019} introduced PolicyLint, an automated tool for analysing privacy policies by identifying contradictory sharing and collection practices using advanced NLP and ontology generation techniques. Applying PolicyLint to a large dataset of privacy policies from Google Play, they found a significant occurrence of logical contradictions and narrowing definitions, highlighting the issue of misleading statements. \citet{Damiano-T2020} used NLP techniques and supervised ML approaches to develop an AI assistant to classify and check the completeness of privacy policies. A case study of applying this tool to 24 privacy policies discovered 45 out of 47 incomplete issues against the GDPR. Recently, \citet{Tang-P2024} proposed a comprehensive GDPR taxonomy and developed a corpus of labelled privacy policies with hierarchical information. Their extensive efforts in evaluating GDPR concept classifiers aimed to enhance the accuracy and reliability of automated GDPR compliance analysis.

\subsubsection{LLM-based Approaches}
\label{subsubsec:privacy_policy_llm}

In contrast to traditional ML and NLP-based methodologies, which often require extensive efforts to label data or rely on existing annotated datasets for supervised learning, emerging LLMs have proven to be powerful and efficient for privacy policy analysis, particularly in classifying data practices and extracting valuable insights without requiring extensive manual annotations or training. Various prompt-based methods have been explored to help human readers navigate complex legal texts in privacy policies. \citet{Tang-C2023} developed PolicyGPT, which leveraged LLMs using a ``prefix prompt'' to perform categorical classification and definition tasks, achieving average Macro F1 scores of 97\% and 87\% for classification tasks using GPT4 on two privacy policy datasets, respectively. It significantly outperformed past ML models. Similarly, \citet{Salvi-RC2024} introduced PrivacyChat, a prompt-engineering-based system using an LLM (GPT-3.5) to improve consumers' comprehension of privacy policies. More recently, \citet{Goknil-A2024} proposed PAPEL, a framework that leverages the capabilities of LLMs through prompt engineering to analyse and convert the critical aspects of privacy policies into user-friendly summaries. Their findings demonstrated that various LLMs, including LLaMA and GPT-3.5/GPT-4, can achieve robust performance in privacy policy analysis tasks. Slightly differently, \citet{Rodriguez-D2024} demonstrated the effectiveness and accuracy of LLM-based solutions for automating privacy policy analysis. They further proposed a set of specific configurations for ChatGPT and argued that such a solution should be considered a replacement for traditional NLP techniques for automated privacy policy processing. More recently, \citet{Chen-Y2025} developed LLM-based privacy policy concept classifiers using both prompt engineering and LoRA (low-rank adaptation) fine-tuning techniques, and the results show that such classifiers can outperform other state-of-the-art classifiers using different LLMs (including GPT-3.5, Llama 3 8B, Qwen1.5 7B). To explore LLM explainability in terms of completeness, logicality, and comprehensibility, they also applied prompt engineering to generate explanations for the classification results. Evaluation by three human annotators indicated that LLMs can provide clear and understandable justifications. Similarly, \citet{Mori-K2025} conducted a study to evaluate LLMs' capabilities to understand privacy policies compared with real human users. The results revealed that LLMs could achieve an accuracy of 85.2\% of comprehension levels, while the correct answer rate of human users was only 63\%, demonstrating the potential of using LLMs to replace user studies for the evaluation of privacy policies. Apart from benefiting ordinary users, LLMs have also been applied to discover legal and reputational risks by detecting policy violations within enterprises' regulatory and operational frameworks. A recent work by \cite{Rachmil-O2025} proposes a training-free method that consider policy violation detection as an out-of-distribution detection problem and adopts whitening techniques. Their method requires policy text and a small number of examples, yet achieves state-of-the-art performance compared with other LLM-as-a-judge and fine-tuning approaches.  

Similar to our work, \citet{Xie-Q2025} conducted a study of using the Llama-3.1-70B-Instruct to support automated analysis of privacy policies, which involves two main tasks. One is the assessment to what extent a given privacy policy segment covers the required content of a specific clause. Another one is to assess personal information practices in privacy policies, specifically, focusing on categories of personal information collects and shares, the purposes of data collection and sharing, and the third-party recipients. More recently, and in close relation to our work, \citet{Yang-M2025} developed a framework that performs a series of operations, including privacy policy text segmentation, paragraph-level classification, sentence-level classification, privacy attribute mapping, relationship extraction, and graph generation. Unlike our approach, the key component of their framework for reconstructing attributes and relationship graphs relies on the OPP-115 dataset as training and testing data, applying knowledge distillation techniques involving GPT models and BERT. Their graph generation is implemented using Neo4j, allowing data flows visualisation and interactive query features. However, the reliance solely on OPP-115 labels may not be suitable for modern privacy policies. Different from past studies, our work leverages taxonomies and findings from multiple existing studies to build a knowledge base, aiming to facilitate LLM-driven privacy policy analysis from the perspective of data flows, specifically, what personal data is collected, with whom it is shared, and for what purposes.

\subsubsection{Data Flow Analysis}

Although LLMs have not been utilised to analyse data flows, such analyses has gained significant attention because they can greatly facilitate the communication of complex concepts in a more accessible and engaging manner~\cite{Karl-2024}. An early effort in this domain was PoliCheck, proposed by \citet{Andow-B2020}, who developed an entity-sensitive flow-to-policy consistency model to identify contradictions between the data flows described in mobile app privacy policies and the actual data handling practices of the apps. Similarly, \citet{Cui-H2023} introduced PoliGraph, a knowledge graph designed to represent data flows, along with an NLP-based tool, PoliGraph-ER, for the automated extraction of data flows from privacy policy texts. Their evaluation demonstrated that PoliGraph-ER outperformed PoliCheck in identifying inconsistencies between stated and actual data flows.

\citet{Yuan-H2023} also conducted a case study of extracting data flows from the privacy policy of Booking.com, showing significant challenges in manually reconstructing data flows to fully capture the data-sharing landscape. This emphasises the need for more advanced and automated approaches to extracting and analysing data flows in privacy policies. In our prior work~\cite{Yuan-H2024}, we employed GPT-4 as an example LLM to analyse the privacy policies of selected car brands, aiming to produce data entity relationships that capture the types of data shared, the intended purposes of data sharing, and the recipients of the shared data. This was part of a broader effort to reconstruct a vehicle-centric data ecosystem. While the primary focus of that study was not the utilisation of LLMs in privacy policy analysis, it nevertheless offered a glimpse into LLMs' potential and directly motivated the follow-up work presented in this paper.

\subsection{Identified Research Gaps}

Although LLMs have been used for privacy policy analysis, as reviewed in Section~\ref{subsubsec:privacy_policy_llm}, they have not yet been specifically applied to facilitate the extraction of comprehensive data flows and construction of data flow graphs. One possible reason is the well-documented issue of LLM hallucination~\cite{Huang-L2024} that can compromise accuracy and produce unreliable results. Hallucinations, particularly those originating from pre-training data, can introduce misinformation, biases, and knowledge gaps~\cite{Huang-L2024}, making it challenging to extract reliable information about data flows in privacy policy texts. Additionally, privacy policies show significant linguistic variability, with different organisations using different terminologies and structures to describe similar concepts, making it more complicated to conduct comparison analyses across privacy policies.

A promising approach to mitigate these challenges is retrieval-augmented generation (RAG)~\cite{Lewis-P2020}, which can effectively minimise hallucinations caused by knowledge gaps while preserving the generative capabilities of LLMs. The RAG approach augments LLMs with external knowledge sources by converting both knowledge base elements and users' queries into numerical representations called embeddings. This allows semantic similarity searches to retrieve the most relevant information from the knowledge base, which is then provided to the LLM alongside the original query. As a result, it enhances the accuracy and credibility of the generative outputs of LLMs.

One key advantage of introducing RAG is that LLMs do not need to be retrained for task-specific applications, making them more efficient and adaptable~\cite{Gao-Y2023}. In addition, considering the diverse linguistic variability shown in different privacy policies, adopting RAG can potentially constrain and unify the generative outputs of LLMs, enabling consistent formatting for more effective and comparative analysis and visualisation.

However, to the best of our knowledge, no past studies have utilised LLMs with RAG to conduct privacy policy analysis for extracting structured data flows in a comprehensive manner. In addition, while prior research has examined privacy policies either manually or with other AI-based methods, a systematic framework for automated processing of privacy policies to extract and visualise data flows remains largely unexplored. 

\section{The Proposed Framework: \SysName}
\label{sec:framework}

\begin{figure*}[!htb]
\centering
\includegraphics[width=\linewidth]{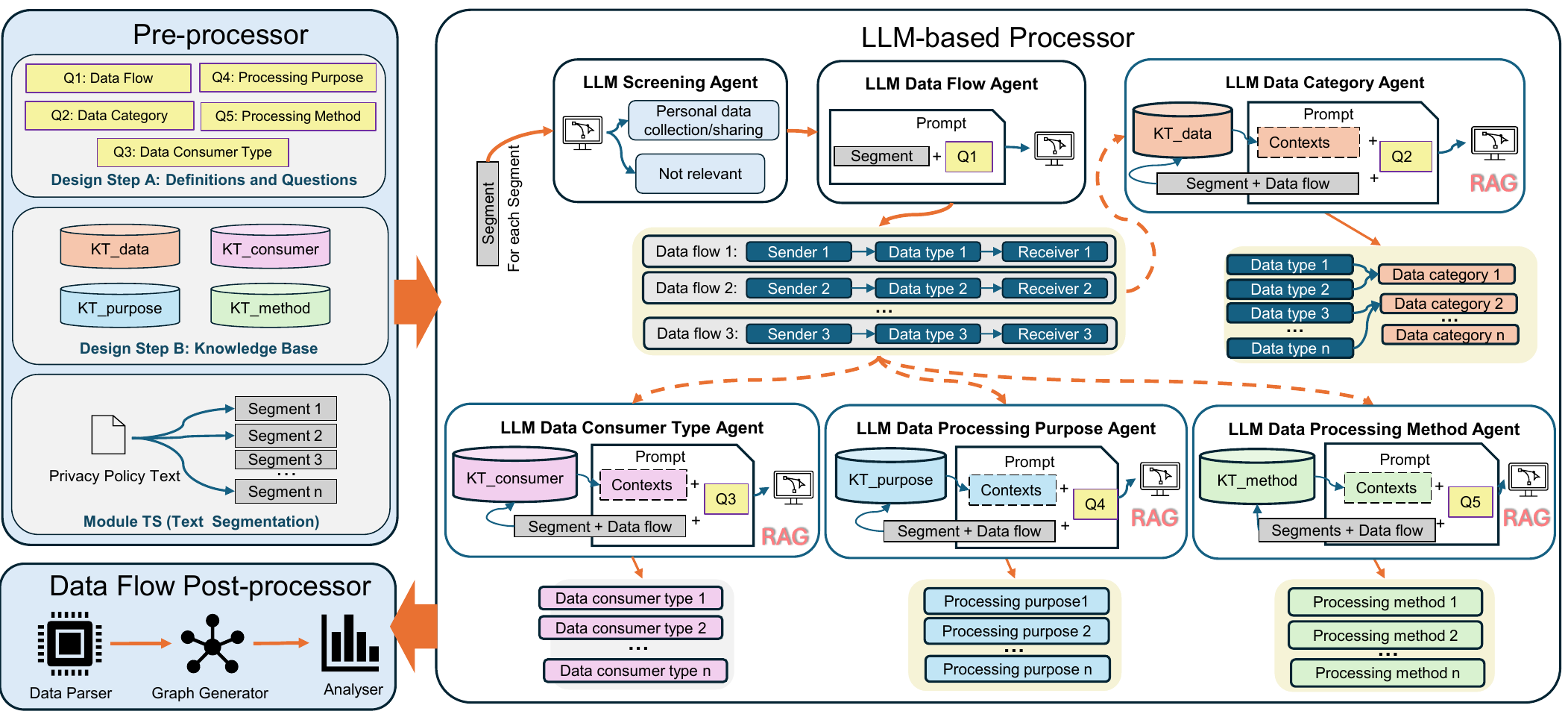}
\caption{\SysName\ architecture}
\label{fig:framework}
\end{figure*}

In this section, we introduce \SysName\ for extracting data flows from privacy policies, constructing and analysing data flow graphs using LLMs and RAG. As illustrated in Figure~\ref{fig:framework}, it consists of three main components: a pre-processor, an LLM-based processor, and a data flow post-processor.\footnote{The source code can be accessed from~\url{https://github.com/hyyuan/LADFA}}.
The pre-processor is responsible for 1) defining concepts to facilitate data flow extraction; 2) constructing a knowledge base; and 3) converting the input privacy policy into machine-readable segments. The LLM-based processor employs a hybrid approach that combines prompt chaining with LLM agents. As shown in Figure~\ref{fig:framework}, each text segment is processed through multiple sequential sub-components, with each sub-component focusing on addressing a specific task. Dashed lines in the figure indicate the flow of the prompt chain, while the highlighted `RAG' texts denote instances where the LLM agent needs dynamically access the knowledge base, retrieve relevant information, and augment the generative output. The data flow post-processor consists of a data parser, a graph generator, and an analyser, to collectively transform, structure, visualise, and interpret the extracted data flows from the LLM-based processor. Note that it is possible to leverage LLMs for the pre- and post-processors as well, but this paper focuses on applying LLMs to the analyser because it is the most complicated part of the whole pipeline so can benefit the most from the use of LLMs. 

\subsection{Pre-processor}

The pre-processor consists of three sub-components: two design steps (A for preparing key questions and B for constructing relevant domain knowledge bases) and one module (TS for text segmentation). Design steps A and B can be processed once and reused thereafter, and Module TS operates dynamically.

\subsubsection{Design Step A: Preparing Key Definitions and Key Questions}
\label{subsubsec:pre-processor_DesignStepA_questions}

As reported in previous studies, key parameters in the CI theory that can affect users' privacy expectations and lead to different privacy comprehensions~\cite{Martin-K2016}. In this work, we aim to analyse privacy policies and extract comprehensive data flows through the lens of CI theory, adopting its parameters with slight modifications to better align with terminologies that have been used in existing studies~\cite{Wilson-S2016, Belen-Saglam-R2022, Bhatia-B2017, Bui-D2021a, Yuan-H2023, Yuan-H2024} and data protection regulations such as the EU/UK GDPR. These refined definitions are used consistently throughout the paper.

One key parameter in the CI theory is `actors', which consists of `subject of information', `the sender of information', and `recipient of information'. Here, we refer the latter twos as \emph{data sender} and \emph{data receiver}, respectively. The `subject of information' in this study is the data subject defined in the GDPR, who are usually the consumer/reader of a privacy policy (but not necessarily so if the reader is taking actions on behalf of other data subjects, e.g., parents managing their children's data). Note that the data subject is not always the data sender, e.g., when an online service shares a user's data with a third-party, the data sender is the online service.

As mentioned earlier, another key parameter of CI theory is `attributes' (i.e., types of information). In the context of this work, attributes are the types of personal information passed from \emph{data sender} to \emph{data receiver}. As reported in \cite{Belen-Saglam-R2022}, personal information can be categorised hierarchically, for instance, `personal identification information' is the top level category, where `contact information' is a subcategory of `personal identification information', and `phone number' and `email address' are subcategories of `contact information'. For consistency and readability, we use \emph{data type} to denote the type of personal information described in privacy policies. In such a hierarchy structure, \emph{data type} may appear as a leaf node (e.g., name or email address), a root node (e.g., personal data), or a mid-level node (e.g., demographic data or location data), depending on the writing style of a privacy policy. In summary, these forms the \emph{data flow} concept discussed in this paper. Specifically \emph{data sender}$\rightarrow$\emph{data type}$\rightarrow$\emph{data receiver} indicates the how a specific type of data flows from one party to another party. 

Furthermore, to align variations of \emph{data types} appeared in privacy policies written in different styles and to support consistent comparative analyses, we introduce the concept of \emph{data category}, aiming to further process and classify \emph{data types} based on a customised  and simplified typology. Existing studies such as~\cite{Belen-Saglam-R2022, Wilson-S2016} and regulations such as the GDPR have categorised personal information, often through complex, multi-layered hierarchical structures. In this study, we propose a simplified three-level typology, intended solely to facilitate and demonstrate the effectiveness and usefulness of the proposed framework. The root-level node of the proposed typology is `personal data', then to differentiate nodes at remaining levels, we refer to a mid-level node as a `\emph{data category}' and a leaf node as a `\emph{data type}'. The term `\emph{data category}' is aligned with the GDPR's notion of `special categories of personal data'. More details of developing the proposed typology are provided in the paragraph ``Knowledge Typology for Classifying Data Categories'' later in this section. Moreover, we introduce \emph{data consumer type}, \emph{data processing purpose}, and \emph{data processing method}, aiming to cover the broad scope the `transmission principles' of the CI theory. 

The concept of \emph{data consumer type} is defined to capture the role of data consumer given a data flow. In line with the GDPR's definition, a data consumer maybe a (data) controller or a (data) processor. In this study, if the data consumer is a controller, which typically owns the privacy policy, its type is considered as first-party. If the data consumer is a processor, its type would depend on different contexts: 1) if it operates with the same organisation, it is considered as first-party; 2) if it is an external entity such as third-party service provider, the data consumer type is regarded as third-party. More detailed description can be found in the paragraph ``Knowledge Typology for Identifying Data Consumer Type''.

The concept of \emph{data processing purpose} is introduced to capture the intended use or objective behind the data processing. While the concept is related to the lawful bases (conditions) defined in GDPR, we did not rely on the regulation to derive the \emph{data processing purpose} applied in this study. Instead, we referred to existing studies~\cite{Wilson-S2016,Bhatia-B2017,Bui-D2021a} that examines privacy policies, from which we established a set of purposes tailored to this study. More details can be found in the paragraph ``Knowledge Typology for Identifying Data Processing Purpose''.  

The \emph{data processing method} refers to the method in which personal data is processed. Inspired by previous studies~\cite{Wilson-S2016, Yuan-H2023}, two main types of data processing methods are introduced in this study: 1) a method is active when the data is voluntarily entered or generated by the user while interacting with the service covered by the privacy policy; 2) a method is passive when the data is automatically collected and shared without user input, so passive from the user's perspective. More detailed discussion can be found in the paragraph ``Knowledge Typology for Identifying Data Processing Method''. 

To this end, \emph{data consumer type}, \emph{data processing purpose}, and \emph{data processing method} further enhance the context of \emph{data flow} to form the \emph{comprehensive data flow}. Building upon the above definitions, we break down the complex task of extracting \emph{comprehensive data flows} into several subtasks, each of which answers a specific question. In this design step, we formulated a set of five questions.

\begin{itemize}
\item Q1: What are the claimed \emph{data flows} described in the privacy policy text, in which a \emph{data receiver} collects a \emph{data type} from a \emph{data sender} or \emph{data sender} shares a \emph{data type} with a \emph{data receiver}?

\item Q2: For each \emph{data type} within a \emph{data flow} identified in Q1, can it be further processed and classified as a specific \emph{data category}?

\item Q3: For each data flow identified in Q1, what is the \emph{data consumer type}? 

\item Q4: For each data flow identified in Q1, what is the \emph{data processing purpose}? 

\item Q5: For each data flow identified in Q1, what is the \emph{data processing method}?
\end{itemize}

To allow \SysName\ to generate reliable answers to these questions, it is important to define the scope associated with these questions so that LLMs can have better contexts to understand and process the input text. To this end, it is important to establish a domain knowledge base that supports answering these questions and facilitates the construction of customised prompts.

\subsubsection{Design Step B: Constructing The Domain Knowledge Base}
\label{subsubsec:pre-processor_DesignStepB_knowledge_bases}

To address questions set in Section~\ref{subsubsec:pre-processor_DesignStepA_questions}, particularly Q2--Q5, we establish a domain knowledge base composed of four knowledge typologies, each tailored to a specific question. The domain knowledge base is subsequently segmented and encoded into vector representations using an embedding model, then stored in a vector database to support the retrieval phase of the RAG approach. This can also help ground LLMs' behaviours to generate outputs that are more consistent and less variable.

To this end, we adopted definitions and examples from multiple existing studies, including the OPP-115 dataset~\cite{Wilson-S2016}, the personal information taxonomy study~\cite{Belen-Saglam-R2022}, a data processing purposes case study~\cite{Bhatia-B2017}, a study of data-usage purposes in mobile apps~\cite{Bui-D2021a}, and findings from our own previous studies~\cite{Yuan-H2023, Yuan-H2024}, to build the domain knowledge base. For the rest of the paper, the domain knowledge base is referred to as $\text{KB}$, which consists of $\text{KT}_{\text{data}}$, $\text{KT}_{\text{consumer}}$, $\text{KT}_{\text{purpose}}$, and $\text{KT}_{\text{method}}$. They represent knowledge typologies for \emph{data category}, \emph{data consumer type}, \emph{data processing purpose}, and \emph{data processing method}, respectively. Table~\ref{tab:kt_summary} summarises how knowledge typologies were derived using existing studies. In the remainder of this section, we describe how each of these studies contribute to the construction of the knowledge typologies in more details. 

\begin{table}[!htb]
\centering
\caption{Summary of how multiple sources were used to construct the four knowledge typologies}
\label{tab:kt_summary}
\begin{tabular}{lcccc}
\toprule
\textbf{Source} & $\text{KT}_{\text{data}}$ & $\text{KT}_{\text{consumer}}$ & $\text{KT}_{\text{purpose}}$ & $\text{KT}_{\text{method}}$\\
\midrule
OPP-115~\cite{Wilson-S2016} & \ding{51} & \ding{51} & \ding{51} & \ding{51}\\
Personal information taxonomy~\cite{Belen-Saglam-R2022} & \ding{51} & & &\\
data processing purposes case study~\cite{Bhatia-B2017} & & & \ding{51} &\\
Data-usage purposes in mobile apps~\cite{Bui-D2021a} & & & \ding{51} &\\
Data flow reconstruction using a privacy policy\cite{Yuan-H2023} & & \ding{51} & \ding{51} & \ding{51}\\
Vehicle-centric data ecosystem~\cite{Yuan-H2024} & \ding{51} & \ding{51} & &\\
\bottomrule
\end{tabular}
\end{table}

\paragraph{Knowledge Typology for Classifying Data Categories}

In this part, we introduce how we constructed $KT_{data}$ for processing and classifying data categories to address Q2. As mentioned earlier, we introduce three-level typology, with the root‑level represented by a single node, `personal data'. Our focus is on defining the remaining levels. We use the OPP-115 dataset~\cite{Wilson-S2016} as a baseline to construct the knowledge typology, and then further refine and expand it using the personal data taxonomies~\cite{Belen-Saglam-R2022} and findings from the work~\cite{Yuan-H2024}.

OPP-115 dataset is a collection of 115 website privacy policies with manual annotations of 23,000 fine-grained data practices~\cite{Wilson-S2016}. It was released in 2016 and has become a widely used resource for privacy policy research. The dataset defines ten data practice categories, each accompanied by detailed descriptions and illustrative examples. The OPP-115 dataset cover personal data in 16 main categories, including \emph{Finance}, \emph{Health}, \emph{Contact}, \emph{Location}, \emph{Demographic}, \emph{Personal Identifier}, \emph{User Online Activities}, \emph{User Profile}, \emph{IP Addresses and Device IDs}, \emph{Cookies and Tracking Elements}, \emph{Computer Information}, \emph{Survey Data}, \emph{Generic Personal Information}, \emph{Other}, and \emph{Unspecified}.

The work conducted by~\citet{Belen-Saglam-R2022} investigates how personal data evolved and was perceived across different domains. They analysed data from multiple sources, including governmental legislation/regulations, privacy policies of applications, and academic research articles, and produced a series of hierarchical personal information taxonomies (see Table~\ref{tab:taxonomy} for a top-level overview). As shown in Table~\ref{tab:taxonomy}, some data categories identified for different domains in \cite{Belen-Saglam-R2022} are consistently identified from multiple (i.e., at least three) data sources, including \emph{Demographic}, \emph{Personal Identification Information}, \emph{Financial Information}, \emph{Health Information}, \emph{Criminal Records/Court Judgements}, \emph{Sex Life \& Sexual Orientation}, and \emph{Communication Data} categories, suggesting these represent common data categories independent of domain context. Among these categories defined in the personal information taxonomies, the first four categories are present in the OPP-115 dataset, but using slightly different terminologies. For consistency and simplification, \emph{Demographic}, \emph{Personal Identity Identifier}, \emph{Finance}, and \emph{Health} are considered as mid-level nodes of $\text{KT}_{\text{data}}$. 

\begin{table}[!htb]
\resizebox{\linewidth}{!}{ %
\begin{threeparttable}
\caption{First order system of categorisation of personal information taxonomies from different data sources~\cite{Belen-Saglam-R2022}}
\label{tab:taxonomy}
\centering
\begin{tabular}{lccccc}
\toprule
\textbf{Categories} & \textbf{Government} & \textbf{App (H)} & \textbf{App (F)} & \textbf{Academic paper (H)} & \textbf{Academic paper (F)}\\
\midrule
Demographic & \ding{51} & \ding{51} & \ding{51} & \ding{51} & \ding{51}\\
Personal Identification Information & \ding{51} & \ding{51} & \ding{51} & \ding{51} & \ding{51}\\ 
Financial Information & \ding{51} & \ding{53} & \ding{51} & \ding{53} & \ding{51}\\
Health Information & \ding{51} & \ding{51} & \ding{53} & \ding{51} & \ding{53}\\
Judicial Data & \ding{51} & \ding{53} & \ding{53} & \ding{53} & \ding{53}\\
Criminal Records/Court Judgements & \ding{53} & \ding{51} & \ding{51} & \ding{51} & \ding{51}\\
Sex Life \& Sexual Orientation & \ding{53} & \ding{51} & \ding{51} & \ding{51} & \ding{51}\\
Technical Device Information & \ding{53} & \ding{51} & \ding{53} & \ding{51} & \ding{53}\\
Communication Data & \ding{53} & \ding{51} & \ding{51} & \ding{51} & \ding{51}\\
Property/Assets Information & \ding{53} & \ding{53} & \ding{51} & \ding{53} & \ding{51}\\
Security Data & \ding{53} & \ding{53} & \ding{53} & \ding{51} & \ding{53}\\
\bottomrule
\end{tabular}
\begin{tablenotes}
\small
\item \ding{51}: The data category is identified from the corresponding data source.
\item \ding{53}: The data category is not identified from the corresponding data source.
\item H: Health, F: Finance
\end{tablenotes}
\end{threeparttable}
}
\end{table}

In addition, \emph{Communication Data}, as defined in~\cite{Belen-Saglam-R2022}, overlaps with several OPP-115 data categories, including \emph{User Online Activities}, \emph{User Profile}, \emph{IP Addresses and Device IDs}, and \emph{Cookies and Tracking Elements}. To avoid redundancy, \emph{IP Addresses}, \emph{Device IDs}, and \emph{Cookies and Tracking Elements} are merged into one single category, \emph{Online Identifier}. Additionally, \emph{Device Information} was included to replace \emph{Computer Information}, aiming to cover a broader range of modern devices (e.g., smartphones, tablets). Furthermore, because the \emph{Security Data} category in~\cite{Belen-Saglam-R2022} includes biometric attributes, which are absent in the OPP-115 dataset, \emph{Biometric Information} was incorporated as another mid-level node of in $\text{KT}_{\text{data}}$.

Moreover, our previous study~\cite{Yuan-H2024} identified governmental bodies as important parts in vehicle-centric data-sharing ecosystems, particularly concerning data collection and sharing involving legal authorities such as courts and law enforcement agencies. This closely aligns with the \emph{Criminal Records/Court Judgements} category listed in Table~\ref{tab:taxonomy}, which is therefore considered as another mid-level node of $\text{KT}_{\text{data}}$.

Following the same practice of refining and merging results using different sources~\cite{Wilson-S2016, Belen-Saglam-R2022, Yuan-H2024}, $\text{KT}_{\text{data}}$ includes the following data categories (i.e., mid-level nodes: \emph{Demographics}, \emph{Contact}, \emph{Finance}, \emph{Health}, \emph{Location}, \emph{Personal Identity Identifier}, \emph{Online Identifier}, \emph{Device Information}, \emph{Biometric Information}, \emph{User Online Activities}, \emph{User Profile}, \emph{Criminal Records/Court Judgements}, \emph{Generic Personal Information}, \emph{Survey data}, \emph{Other}, and \emph{Unspecified}. Apart from \emph{Other} and \emph{Unspecified}, each data category is accompanied by a text description and a list of data types (i.e., leaf nodes) that serve as illustrative examples belonging to the corresponding data category. It is worth noting that these leaf nodes are mainly derived from~\cite{Wilson-S2016, Belen-Saglam-R2022, Yuan-H2023} and are not exhaustive. An example branch of $\text{KT}_{\text{data}}$ is shown below:

\begin{quote}
\textbf{Root node:} Personal data

\textbf{Mid-level node:} Location

Description: Geo-location information (e.g., a user’s current location) regardless of granularity, which may include exact location, ZIP code, or city‑level data.

\textbf{Leaf nodes:} Location data, Global Positioning System (GPS) location data, Location history, Global System for Mobile communications (GSM) location data, Universal Mobile Telecommunications Service (UMTS) location data.
\end{quote}

It has been well documented that, LLMs can perform better on various tasks with few-shot prompting compared to zero-shot prompting, with few-shot prompting's performance in some cases becoming comparable with fine-tuning approaches~\cite{Brown-TB2020}. Including such descriptive text alongside an array of \emph{data types} can further enhance LLM performance by providing relevant context and examples to support few‑shot prompting. To support this as well as the implementation of RAG, the typology is encoded in JSON format\footnote{Knowledge typologies encoded in JSON can be accessed from~\url{https://osf.io/ab23w/overview?view_only=23e83d260dcf419899585ce868ace61b}}. The same approach is applied to the remaining knowledge typologies presented in the rest of the section.

\paragraph{Knowledge Typology for Identifying Data Consumer Type}

For the knowledge typology to identify data consumer type and address Q3, we applied the same hierarchical structure as $\text{KT}_\text{data}$ to develop $\text{KT}_{\text{consumer}}$. The focus is on defining mid-level and leaf nodes. As introduced in Section~\ref{subsubsec:pre-processor_DesignStepA_questions}, the two main data consumer types are first-party and third-party, which are considered as the mid-level nodes. To define them, we adopted the following definitions from the OPP-115 dataset:
\begin{itemize}
\item First-party collection and use: `\emph{Privacy practice describing data collection or data use by the company/organisation owning the website or mobile app}', and

\item Third-party collection and use: `\emph{Privacy practice describing data sharing with third parties or data collection by third parties. A third-party is a company or organisation other than the first-party company or organisation that owns the website or mobile app}.'
\end{itemize}

To generalise these definitions beyond websites and mobile apps, we slightly change the text description and introduce non-exhaustive lists of first-party and third-party entities as leaf-nodes of $\text{KT}_\text{consumer}$. These leaf-nodes are informed by prior studies on examining privacy policies~\cite{Yuan-H2023, Yuan-H2024}, which identified various frist-party and third-party entities. As an example, the following branch demonstrates the structure of $\text{KT}_\text{consumer}$:

\begin{quote}
\textbf{Root node:} Data consumer type

\textbf{Mid-level node:} First Party

Description: A first party is the entity, such as a website or company, that directly collects and uses personal data from individuals/customers. The company/website/application/service's actual name would be often used as indication of existence of the first party.

\textbf{Leaf nodes:} We, Us, This website, This company, This organisation, Our website, Our company, Our organisation, Our service.
\end{quote}

\paragraph{Knowledge Typology for Identifying Data Processing Purpose}

With respect to Q4, we define $\text{KT}_{\text{purpose}}$) as a two-level typology, with the root-level node represented by a single node capturing the overarching objective of data processing. Then the focus is on developing an array of leaf nodes with corresponding descriptions that specify distinct purposes, derived from existing studies. Few existing studies have specifically focused on examining data processing purposes in privacy policies. \citet{Bhatia-B2017} recruited human annotators to study five privacy policies, producing 218 data purpose annotations. Their analysis identifies six categories of data processing purposes: \emph{Service Purpose}, \emph{Legal Purpose}, \emph{Communication Purpose}, \emph{Protection Purpose}, \emph{Merger Purpose}, and \emph{Vague Purpose}. \citet{Bui-D2021a} developed a hierarchical taxonomy of data usage purposes by applying neural text clustering with contextualised word embeddings to group purpose clauses that have similar meanings in a large policy corpus. The hierarchical taxonomy consists of four high-level purposes and 16 low-level purposes (see Table~\ref{tab:hierarchy_purpose_taxonomy}). The OPP-115 dataset~\cite{Wilson-S2016} includes detailed descriptions of 11 purposes, including \emph{Basic Service/Feature}, \emph{Additional Service/Feature}, \emph{Advertising}, \emph{Marketing}, \emph{Analytics/Research}, \emph{Personalisation/Customisation}, \emph{Service Operation and Security}, \emph{Legal Requirement}, \emph{Merger/Acquisition}, \emph{Other}, and \emph{Unspecified}. These purposes were manually created by three recruited law experts using a top-down approach.

\begin{table}[!htb]
\centering
\caption{The hierarchical taxonomy of data-usage purposes in~\cite{Bui-D2021}}
\label{tab:hierarchy_purpose_taxonomy}
\resizebox{0.4\linewidth}{!}{ %
\begin{tabular}{ll}
\toprule
\textbf{High-level} & \textbf{Low-level}\\
\midrule
Production & Provide service\\
& Improve Service\\
& Personalise Service\\
& Develop Service\\
& Manage Service\\
& Manage Accounts\\
& Process Payments\\
& Security\\
\midrule
Marketing & Customer Communication\\
& Marketing Analytics\\
& Promotion\\
& Provide Ad\\
& Personalise Ad\\
& General Marketing\\
\midrule
Legality & General legality\\
\midrule
Other & Other purposes\\
\bottomrule
\end{tabular}
}
\end{table}

By comparing all the existing sets of purposes, we found that those defined in~\cite{Bhatia-B2017} and \cite{Bui-D2021a} are largely reflected in the OPP-115 definitions. However, certain purposes in the OPP-115 dataset, such as \emph{Merger/Acquisition} and \emph{Analytics/Research}, are not explicitly captured in the hierarchical taxonomy proposed in~\cite{Bui-D2021a}. Moreover, the definitions provided by \citet{Bhatia-B2017} are relatively brief and overlap substantially with those in the other two sources. Additionally, as noted in one or our previous studies~\cite{Yuan-H2023}, various personal data are often collected and shared to enable social media integration with websites/apps. We observed that this purpose is absent from the categories in~\cite{Bhatia-B2017, Bui-D2021, Wilson-S2016}.

Considering all the observations and given the broader yet fine-grained set of data processing purposes in OPP-115, we adopted this set as the primary leaf nodes of $\text{KT}_{\text{purpose}}$ with\emph{Social Media Integration} explicitly added as an additional leaf node in this study. In total, there are 10 leaf nodes of data processing purposes including \emph{Basic Service or Feature}, \emph{Additional Service or Feature}, \emph{Advertising}, \emph{Marketing}, \emph{Analytics or Research}, \emph{Personalisation or Customisation}, \emph{Operational Integrity and Security}, \emph{Legal requirement}, \emph{Merger/Acquisition}, and \emph{Unspecified}. An illustrative branch of $\text{KT}_{\text{purpose}}$ is presented below:

\begin{quote}
\textbf{Root node:} Data processing purpose

\textbf{Leaf node:} Advertising

Description: To show advertisements that are either targeted to the specific user or not targeted, or other general advertising activities.
\end{quote}

It is worth noting that the data processing purposes presented here are not intended to cover all lawful bases (conditions) under the GDPR, nor is the aim to produce a comprehensive dataset for this purpose. Nevertheless, comparing existing categorisations/taxonomies with definitions in the GDPR is an interesting topic for future research, however, it is out of the scope of this work.

\paragraph{Knowledge Typology for Identifying Data Processing Method}

To help address Q5, we mainly use findings from \cite{Yuan-H2023} and definitions in the OPP-115 dataset as references to complete the knowledge typology. Similar to $\text{KT}_{\text{purpose}}$, $\text{KT}_{\text{method}}$ is structured as a two-level typology, with a single root node and a set of leaf nodes with the corresponding descriptions. A manual privacy policy analysis reported in~\cite{Yuan-H2023} identified two primary methods of data processing: explicit and implicit methods. This finding aligns with the definitions in the OPP-115 dataset, which defines explicit, implicit, and unspecified data processing modes. In these definitions, explicit refers to cases where users actively provide their data while using the service/website, while implicit refers to data collection and use that occurs passively and automatically without active user involvement. Since the term `explicit' is often associated with consent in the context of the GDPR, to ensure clarity, consistency, and avoid confusion, we decided to use the terms \emph{Active}, \emph{Passive}, and \emph{Unspecified} as the three leaf nodes $\text{KT}_{\text{method}}$. To illustrate, one branch of $\text{KT}_{\text{method}}$ is shown below:

\begin{quote}
\textbf{Root node:} Data processing method

\textbf{Leaf node:} Active

Description: a company or organization gathers information that a user knowingly and intentionally provides. This involves instances where users actively input data, such as filling out a web form, creating an account, making a purchase, or subscribing to a newsletter. Explicit data collection typically requires the user to be fully aware of the information being provided and often involves their direct consent.
\end{quote}

\subsubsection{Module TS for Text Segmentation}
\label{subsubsec:pre-processor_Module_TS}

\begin{figure}[!htb]
\centering
\includegraphics[width=\linewidth]{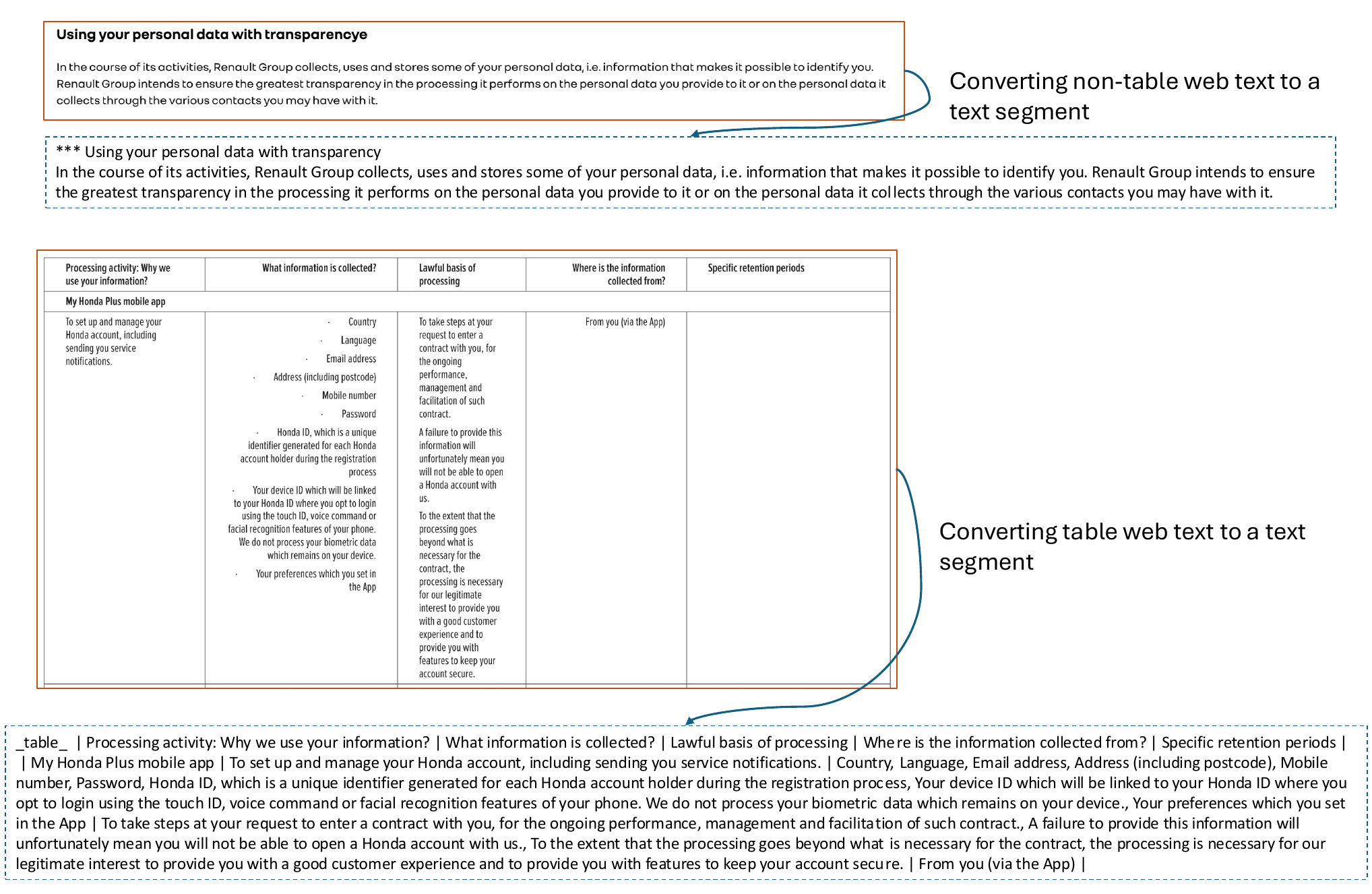}
\caption{Example of converting non-table and table text to text segments}
\label{fig:text_segmentation}
\end{figure}

Several existing tools, such as ASDUS~\cite{Gopinath-M2018} and the \texttt{html2text} Python package~\cite{Savand-A2025}, support automatic text segmentation of HTML documents. However, these tools are either specialised for extracting top-level titles only or struggle with handling table contents and (nested) bullet points. We have observed that many privacy policies frequently use tables and nested bullet points for its content, e.g., for outlining data processing practices. Existing solutions may not effectively convert such HTML content into meaningful text segments for further analysis. Considering this, a customised pipeline utilised the Beautifulsoup4 Python package\footnote{\url{https://pypi.org/project/beautifulsoup4/}} was developed for automatic text segmentation of an HTML page. It consists of the following main steps: \RomanNumber{1}) removing elements associated with heading, footer, style, and scripts (i.e., \verb|<head>|, \verb|<footer>|, \verb|<style>|, \verb|<script>|); \RomanNumber{2}) processing all non-\verb|<table>| HTML elements following Algorithm~\ref{alg:extract_process_text} to produce an initial list of segments; \RomanNumber{3}) extracting table contents with their headings, where each table is one segment; and \RomanNumber{4}) for segments that contain bullet points, identifying the associated heading or paragraph, and merging them with the associated bullet points to form one segment (See Figure~\ref{fig:text_segmentation} in Section~\ref{sec:appendix} for an example).

\begin{algorithm}[!htb]
\centering
\cprotect\caption{The algorithm for extracting and processing non-\verb|<table>| content of an HTML page}
\label{alg:extract_process_text}
\footnotesize
\setlength{\itemsep}{0pt}
\setlength{\parsep}{0pt}
\setlength{\parskip}{0pt}
\begin{minipage}[t]{0.48\textwidth}
\begin{algorithmic}[1]
\State $\text{segments} \gets \varnothing$ \Comment{(Initialise an empty list)}
\ForAll{$\text{element} \in \text{FindAllTags}(\texttt{html})$} 
    \If{element $\in$ \{\texttt{p}, \texttt{h1}, \texttt{h2}, \texttt{h3}, \texttt{h4}, \texttt{h5}, \texttt{li}, \texttt{ol}, \texttt{ul}\}}
        \If{FindParents(element, \{\texttt{ol}, \texttt{ul}\}) $= \varnothing$} 
             \State $\text{segments} = \Call{Process}{\text{element}, \text{segments}}$
        \EndIf
    \EndIf
\EndFor
\end{algorithmic}
\end{minipage}
\hfill
\begin{minipage}[t]{0.48\textwidth}
\begin{algorithmic}[1]
\Function{Process}{element, segments}
    \If{FindParent(element) $\neq$ \texttt{table}}
        \If{element is \texttt{p}}
            \State text $\gets$ ExtractTextWithout<a>(element)
            \If{text $\neq \varnothing$} \State Append text to segments
            \EndIf
        \ElsIf{element in \{\texttt{h1}, \texttt{h2}, \texttt{h3}, \texttt{h4}, \texttt{h5}\}}
            \State text $\gets$ ExtractText(element)
            \If{$\text{text} \neq \varnothing$} \State Append ``*'' + text 
            \EndIf
        \ElsIf{element is \texttt{li} and \texttt{a} not in element}
            \If{element contains no \texttt{a} tags}
                \If{element has nested lists}
                    \State text $\gets$ FlattenNestedLists(element)
                \Else
                    \State text $\gets$ ExtractText(element)
                \EndIf
                \If{text $\neq \varnothing$} \State Append ``$-$'' + text 
                \EndIf
            \EndIf
        \ElsIf{element in \{\texttt{ul}, \texttt{ol}\}}
            \ForAll{child in FindChildren(element)}
                \State segments = \Call{Process}{child, segments}
            \EndFor
        \EndIf
    \EndIf
    \State result $\gets$ Join(segments, END\_OF\_LINE)
    \State \textbf{return} result
\EndFunction
\end{algorithmic}
\end{minipage}
\end{algorithm}

\subsection{LLM-based Processor}

The LLM-based processor adopts a mixed approach that utilises prompt chains and LLM agents. In the context of prompting techniques for generative artificial intelligence (GenAI), an effective prompt engineering technique is to decompose tasks into several subtasks. Prompt chaining refers to sequentially prompting an LLM with a subtask and then using its response as input for the next prompt~\cite{Wu-T2022}. Additionally, RAG plays a crucial role in improving the model outputs by incorporating relevant external knowledge. According to the taxonomies reported in a survey~\cite{Schulhoff-S2024}, RAG systems in the context of GenAI can be classified as agents when considering the retrieval component as an independent tool. As illustrated in Figure~\ref{fig:framework}, the LLM-based processor processes each text segment sequentially through multiple sub-components, with each one responsible for answering one specific question outlined in Section~\ref{subsubsec:pre-processor_DesignStepA_questions}. The dashed lines between the sub-components in Figure~\ref{fig:framework} represent the use of prompt chain techniques, whereas the highlighted `RAG' texts indicate that the corresponding sub-component function as LLM agents.

\subsubsection{LLM Screening Agent}

Assume that a single privacy policy $T$ consists of $n$ text segments, i.e., $T = \{T_i\}_{i=1}^n$. Not all segments are about personal data collection and data sharing practices. The main objective of this sub-component is to filter out irrelevant text segments and pass only the relevant ones to the next sub-component.

\subsubsection{LLM Data Flow Agent}
\label{subsubsec:llm_data_flow_agent}

If a text segment $T_i$ passes the screening phase, it is fed into this sub-component (i.e., LLM data flow agent) to extract data flows. A customised prompt $P_i^{\text{flow}}$ is formulated, as illustrated in Eq.~\eqref{eq:flow_prompting} based on multiple inputs, including the given $T_i$, question Q1 defined in Section~\ref{subsubsec:pre-processor_DesignStepA_questions}, and a set of predefined answering rules $R_{\text{flow}}$. Here, $R_{\text{flow}}$ regulates how an LLM generate outputs and determine their format. Please see Figure~\ref{fig:prompt_dataflow} in Section~\ref{sec:appendix} for a prompt example. 

\begin{equation}
\label{eq:flow_prompting}
P_i^{\text{flow}} = \text{PromptGeneration}\left(T_i, \text{Q1}, R_{\text{flow}}\right)
\end{equation}

Then LLM processes the prompt $P_i^{\text{flow}}$ and generates structured outputs $O_i^{\text{flow}}$ as shown in Eq.~\eqref{eq:llm_flow_output}, where $O_i^{\text{flow}}$ contains $m$ individual data flows, i.e., $O_i^{\text{flow}} = \{F_{i,j}\}_{j=1}^m$.
\begin{equation}
\label{eq:llm_flow_output}
O_i^{\text{flow}} = \text{LLM}_{\text{generate}}\left(P_i^{\text{flow}}\right)
\end{equation}

If there are more than one data sender or data receiver for a identified data flow, multiple data flows will be generated for each unique pair of data sender and data receiver. Each data flow $F_j$ consists of a data sender $\text{DS}_j$, a data type $\text{DT}_j$, and a data receiver $\text{DR}_j$. 
\begin{equation}
F_{i,j} = \left(\text{DS}_{i,j}, \text{DT}_{i,j}, \text{DR}_{i,j}\right), \quad \forall j \in \{1, \ldots, m\}
\end{equation}
If either the data sender or receiver is unknown, we instruct the LLM to leave the corresponding field empty. In such cases, we use the placeholder symbol $\bot$ to represent the missing value, resulting in one of two partial tuples $\left(\bot, \text{DT}_{i,j}, \text{DR}_{i,j}\right)$ or $\left(\text{DS}_{i,j}, \text{DT}_{i,j}, \bot\right)$.

\subsubsection{LLM-RAG Agents}

To further analyse data flows, we develop a set of LLM-RAG agents to identify: 1) data category, 2) data consumer type, 3) data processing purpose, and 4) data processing method. The retrieval phase of the RAG approach involves identifying the corresponding knowledge typology and then retrieve information that is relevant to a given query. Across the RAG and LLM literature, a variety of terms have been used to describe the retrieved component. Some studies refer to the retrieved information as retrieved contexts~\cite{Jin-B2024}, others describe the retrieved text documents as additional contexts~\cite{Lewis-P2020}, while others denote the retrieved snippets as contexts~\cite{Joren-H2025} or contextual backgrounds~\cite{Sahoo-P2025}. For the clarity and simplicity, here we call the retrieved information knowledge contexts.

Each LLM-RAG agent performs a series of similar tasks that can be generalised as the following:

\begin{enumerate}
\item 
As part of the RAG process, the task is to retrieve knowledge contexts that are most relevant to a given text segment and a single data flow from the set of extracted data flows, based on semantic similarity. This task can be represented in Eq.~\eqref{eq:rag_retrieval}. Here, $\text{KCX}_{i,j}$ represents the top-$k$ retrieved knowledge contexts, where the value of $k$ is dynamically determined. The retrieval is carried out using the \verb|VectorIndexRetriver| module of LlamaIndex~\cite{Liu_LlamaIndex_2022}\footnote{LlamaIndex is a data framework specialised for building applications using LLMs} to embed the query and compares it against the stored knowledge embeddings as discussed in Section~\ref{subsubsec:pre-processor_DesignStepB_knowledge_bases}. By default, the similarity is computed via cosine similarity, ensuring that the retrieved knowledge contexts most semantically aligned with the query are prioritised. In practice, we use a semantic score threshold of 0.6 to retrieve knowledge contexts, therefore, only contexts with a score greater than 0.6 are selected. If a single knowledge context meets this requirement, it is solely used in the prompt. If two or more contexts satisfy the threshold, the top two are included. If no context exceeds 0.6, the one with the highest score is used. $\text{KT}_l$ represents the domain knowledge typology selected for the retrieval, where $l \in \{\text{data}, \text{consumer}, \text{method}, \text{purpose}\}$ corresponds to the knowledge typologies defined in Section~\ref{subsubsec:pre-processor_DesignStepB_knowledge_bases}. $F_{i,j}$ represents the extracted data flow described in Section~\ref{subsubsec:llm_data_flow_agent}, where $^*$ suggests that not all LLM-RAG agents take a completed data flow as input in the retrieving process. One example of using only part of a data flow is detailed in the paragraph ``Agent for Identifying Data Category''. In addition, $T_{\text{adj}}$ represents optional adjacent text segments (e.g., one example of using $T_{\text{adj}}$ is detailed in the paragraph ``Agent for Identifying data processing method'' of this sub-subsection).
\begin{equation}
\label{eq:rag_retrieval}
\text{KCX}_{i,j} = \text{Retrieve}\left(\text{KT}_l, F_{i,j}^*, T_i, T_{\text{adj}}\right)
\end{equation}

\item Once the relevant knowledge context(s) is/are retrieved, a customised prompt $P_{i,j}$ is generated, as formulated in Eq.~\eqref{eq:rag_prompt}. The prompt is constructed using multiple inputs, including the original text segment $T_i$, and the retrieved contexts, $\text{KCX}_{i,j}$. Additionally, one question $\text{Q}p$ is selected, where $p \in \{2,\ldots,5\}$ to guide the prompt formulation. Furthermore, $R$ regulates LLM's output format.
\begin{equation}
\label{eq:rag_prompt}
P_{i,j} = \text{PromptGeneration}\left(T_i, \text{KCX}_{i,j}, \text{Q}p, R\right)
\end{equation}

\item The generated prompt $P_{i,j}$ is then passed to the LLM for inference, producing one structured output $O_{i,j}$, as defined in Eq.~\eqref{eq:llm_output}.
\begin{equation}
\label{eq:llm_output}
O_{i,j} = \text{LLM}_{\text{generate}}(P_{i,j})
\end{equation}
\end{enumerate}

To distinguish different LLM-RAG agents, we use different superscripts in the notations of Eqs.~\eqref{eq:rag_retrieval}, \eqref{eq:rag_prompt}, and \eqref{eq:llm_output} in the remaining part of the section.

\paragraph{Agent for Identifying Data Category}

First, this LLM-RAG agent takes the data type $\text{DT}_{i,j}$ of an identified data flow $F_{i,j}$, a text segment $T_i$, and the knowledge typology of data category $\text{KT}_{\text{data}}$, to retrieve knowledge context(s) $\text{KCX}_{i,j}^{\text{data}}$. The retrieving process follows the same formulation as Eq.~\eqref{eq:rag_retrieval}, with the notations substituted according to those introduced above. 

Then, a customised prompt $P_{i,j}^{\text{data}}$ is dynamically augmented using the retrieved knowledge context(s) $\text{KCX}_{i,j}^{\text{data}}$, $R$, along with the data categorisation question Q2. This can be formalised based on Eq.~\eqref{eq:rag_prompt} accordingly. See Figure~\ref{fig:prompt_data_cateogry} in Section~\ref{sec:appendix} for an example of generated prompt. Finally, the LLM takes the customised prompt $P_{i,j}^{\text{data}}$ and generates one structured output for the data category (i.e., $O_{i,j}^{\text{data}}$).

\paragraph{Agent for Identifying Data Consumer Type}

The objective of this agent is to determine whether the data consumer type of a given data flow $F_j$ is a first-party, third-party, or undefined. Following the same RAG approach, the context retriever takes $F_{i,j}$, $T_i$, and knowledge typology $\text{KT}_{\text{consumer}}$ to retrieve the relevant knowledge context(s) $\text{KCX}_{i,j}^{\text{consumer}}$. A customised prompt $P_j^{\text{consumer}}$ is then constructed and passed to an LLM to identify a single data consumer type (i.e., $O_{i,j}^{\text{consumer}}$).

\paragraph{Agent for Identifying Data Processing Purpose}

This agent determines the data processing purpose for a given data flow $F_{i,j}$. Using the same approach, the associated domain knowledge typology $\text{KT}_{\text{purpose}}$ and a given text segment $T_i$ are used to retrieve the knowledge context(s) $\text{KCX}_{i,j}^{\text{purpose}}$, which is then used to facilitate the prompt generation. The output of this LLM-RAG agent is $O_{i,j}^{\text{purpose}}$, which identifies the data processing purpose for the given data flow.

\paragraph{Agent for Identifying Data Processing Method}

As described in Section~\ref{subsubsec:pre-processor_DesignStepB_knowledge_bases}, the data processing method can be categorised as active, passive, or unspecified. Unlike other agents described above, determining the data processing method is more challenging, as it often depends on the surrounding textual context. Preliminary experiments indicated that incorporating adjacent text segments, i.e., the previous segment $T_{i-1}$ and the next segment $T_{i+1}$, can help improve classification accuracy. Hence, the retrieval task can be reformulated with minor changes to Equation~\ref{eq:rag_retrieval} and represented as follows:
\begin{equation}
\label{eq:data_retrieve_method}
\text{KCX}_{i,j}^{\text{method}} = \text{Retrieve}\left(F_{i,j}, T_{i-1}, T_i, T_{i+1}, \text{KT}_{\text{method}}\right)
\end{equation}
Similar to the above two agents, only the retrieved knowledge context with highest semantic similarity is used. The prompt and output generation processes remained the same as those in the previous sections. The final output of this agent is denoted as $O_{i,j}^{\text{method}}$.

In summary, by processing $n$ privacy policy text segments, the LLM-based processor generates an $n$ sets of outputs $O$. Each set $O_i$ includes $m$ sets of outputs, where each one of these outputs consists of the identified data flow, the corresponding data category, data consumer type, data processing purpose, and data processing method. These can be represented formally in Eq.~\eqref{eq:final_output}.

\begin{equation}
\label{eq:final_output}
 O = \left\{O_i= \left\{F_{i,j}, O_{i,j}^{\text{data}}, O_{i,j}^{\text{consumer}}, O_{i,j}^{\text{purpose}}, O_{i,j}^{\text{method}}\right\}_{j=1}^m\right\}_{i=1}^n
\end{equation}

\subsubsection{Further Implementation Details}
\label{subsubsection:implementation}

As mentioned earlier, embedding, indexing, and retrieval were developed using LlamaIndex~\cite{Liu_LlamaIndex_2022}. Specifically, the embedding uses the \verb|BAAI/bge-small-en-v1.5| embedding model~\cite{Xiao-S2024} to convert textual documents into vector representations that are suitable for retrieval and analysis. The execution of LLM inference is supported by Groq\footnote{\url{https://groq.com/}}, which provides language processing units (LPUs) optimised for accelerating AI inference tasks. The Groq API is compatible with LlamaIndex and supports various open-source LLMs, facilitating seamless integration into \SysName. Different LLMs were tested to find the best ones(s). The screening agent and the data flow extraction agent utilise the \verb|llama-3.3-70b-versatile| model\footnote{\url{https://console.groq.com/docs/model/llama-3.3-70b-versatile}}, while the agent responsible for identifying data categories uses \verb|llama3-70b-8192| model\footnote{\url{https://console.groq.com/docs/model/llama3-70b-8192}, this model, with 70B parameters, was the most up-to-date large model available when we conducted the experiment. It is deprecated now in the Groq platform.}. For computational efficiency, a lightweight model \verb|llama-3.1-8b-instant|\footnote{\url{https://console.groq.com/docs/model/llama-3.1-8b-instant}} was used for tasks involving smaller knowledge bases, i.e., $\text{KT}_{\text{consumer}}$, $\text{KT}_{\text{purpose}}$, $\text{KT}_{\text{method}}$.

Finally, prompts were construction by following the format required by the Groq Chat Completions API to ensure structured and effective interactions with the selected LLMs. In addition, for all LLMs, we set both \verb|top_p| and \verb|temperature| to be 0.5, aiming to reduce sampling variance and constrain token selection, aiming to produce balanced outputs that favour consistency over creativity.

\subsection{Data Flow Post-processor}

To obtain insights from the outputs generated by the LLM-based processor, this component comprises three key sub-components: a data parser, a graph generator, and an analyser.

\subsubsection{Data Parser}
\label{subsubsec:post-processor_data_parser}

The data parser is responsible for data cleaning and disambiguation, ensuring that the raw outputs from the LLM-based processor are refined for subsequent graph generation and analysis. Despite providing detailed instructions to LLMs for structuring their output, inconsistencies and ambiguities were observed in the generated data. To address these challenges, additional processing steps are required.

\paragraph{Duplicate Data Entries}
Redundant entries for the same data (e.g., data sender, data receiver, and data type) often occur due to variations in plural and singular forms or inconsistencies in abbreviations and their corresponding full representations. For instance, `customers' and `customer' could be extracted as two distinct data receivers, or `vehicle identification number' and `vin' might be recognised as separate data types. To mitigate this issue, rule-based approaches utilising the Python package inflect\footnote{\url{https://pypi.org/project/inflect/}} and Python's re module\footnote{\url{https://docs.python.org/3/library/re.html}} are applied to standardise entity representations.
    
\paragraph{Misclassification of Data Consumer Types}
In general, LLMs may incorrectly classify the data consumer type given a data flow, where the data consumer type that should be classified as a first-party is instead treated as a third-party, or vice versa. For instance, in our case study of OEM privacy policies (Section~\ref{sec:case_study}), we observed cases where an OEM's mobile app was identified as a data receiver to collect and process personal data, and the corresponding data consumer type was incorrectly categorised as a third party rather than as a first party. To address this issue and support more in-depth analysis, the data parser analyses the data sender and the data receiver collectively and assigns appropriate attributes through a two-step process:
\begin{enumerate}
\item A list of first-party keywords, including `we', `us', `app', and `website', were generated. Then, the NLP package spaCy\footnote{\url{https://spacy.io/}} is used to extract the root and the possessive modifier (if present) from a given text (i.e., data receiver/sender). If the extracted root text matches any of the predefined first-party keywords or the name of the organisation owning the privacy policy, additionally, if a possessive modifier is present and also matches a keyword from the list or the name of the organisation, the attribute `first-party' is assigned to the associated data receiver/sender. Otherwise, the attribute `third-party' is designated.

\item We introduce the concept of a `user-party' as another attribute, which applies specifically to data senders. Following a similar NLP approach mentioned above, but with a different set of keywords including `you', `user', and `customer', we assigned `user-party' attributes to relevant data senders.
\end{enumerate}

By implementing this approach, the data parser enriches the semantic representations of both data senders and receivers. In turn, this enable a more refined classification of data consumer type of a given data flow. As depicted in Table~\ref{tab:data_flows}, data flows with first-party data consumer type can be represented by three distinct cases, while data flows with third-party data consumer type follow the similar pattern. For the clarity and consistency, we refer to these as first-party data flows and third-party data flows for the rest of this paper.  

\begin{table}[!htb]
\centering
\caption{Data flows cases under different data consumer types}
\begin{tabular}{l}
\toprule
\textbf{First-party data flow}:\\
(User-party) data sender $\rightarrow$ data type $\rightarrow$ (First-party) data receiver\\
(First-party) data sender $\rightarrow$ data type $\rightarrow$ (First-party) data receiver\\
(Third-party) data sender $\rightarrow$ data type $\rightarrow$ (First-party) data receiver\\
\midrule
\textbf{Third-party data flow}:\\
(User-party) data sender $\rightarrow$ data type $\rightarrow$ (Third-party) data receiver\\
(First-party) data sender $\rightarrow$ data type $\rightarrow$ (Third-party) data receiver\\
(Third-party) data sender $\rightarrow$ data type $\rightarrow$ (Third-party) data receiver\\
\midrule
\textbf{Incomplete data flows}:\\
data sender $\rightarrow$ data type $\rightarrow$ ?\\
? $\rightarrow$ data type $\rightarrow$ data receiver\\
? $\rightarrow$ data type $\rightarrow$ ?\\
\bottomrule
\end{tabular}
\label{tab:data_flows}
\end{table}

Moreover, previous research~\cite{Yuan-H2023, Karl-2024} indicated that privacy policies often fail to clearly specify how data are collected, to what extent, and with whom they are shared. To capture such insights, we instructed the LLM data flow agent, as described in Section~\ref{subsubsec:llm_data_flow_agent}, to leave the data sender or data receiver as unknown entities if the text segment does not explicitly specify whose data are processed or who is responsible for the data processing, respectively. We define such cases as incomplete data flows. Incomplete data flows consists of three distinct cases, as illustrated in Table~\ref{tab:data_flows}, where a question mark represents an unknown entity. These enhancements allow us to conduct a more granular analysis and adds interpretability to the follow-up analysis.

\subsubsection{Graph Generator}

The graph generator is responsible for converting the processed data into a graph-based model, which can be formalised as a directed graph describing how data flows between different entities (i.e., data sender, data type, and data receiver). The graph can be formally described as $\mathcal{G} = (\mathcal{V}, \mathcal{E})$, where $\mathcal{V}=\{\mathcal{V}_i\}_{i=1}^M$ and $\mathcal{E}=\{\mathcal{E}_j\}_{j=1}^N$ represent a set of $M$ nodes and a set of $N$ edges, respectively. Each node $\mathcal{V}_i$ represents one entity, and different entity types with different attributes are encoded with different colours. Edges in $\mathcal{G}$ can be visualised in different colours, each representing a different data processing purpose. The implementation of the graph generator is supported by the Python packages networkX\footnote{\url{https://networkx.org/}} and pyvis\footnote{\url{https://pyvis.readthedocs.io/}}.

\subsubsection{Analyser}

The analyser extracts and reports various statistics to facilitate a deeper understanding of the processed privacy policy data. It mainly reports results of analyses based on 1) the network graph constructed in the Graph Generator, which represents the relationships and data flows between different entities, and 2) outputs of the LLM-based processor. This allows comparisons across different privacy policies, identification of common patterns, and insights discovery. Further details on the specific statistics and their implications are presented in the next section.

\section{A Case Study for the Automotive Industry}
\label{sec:case_study}

Advancements in AI, the Internet of Things (IoT), and 5G/6G technologies have revolutionised the automotive industry, enabling connected vehicle services and autonomous driving through the integration of electronic control units (ECUs) and sensors that collect, process, and share vast amounts of vehicle and personal data. The extensive scale of data collection and sharing within the modern vehicle ecosystem~\cite{Li-Y2023, Yuan-H2024} raises significant privacy and security risks for consumers. A recent review published by Mozilla~\cite{Caltrider-J2023} claimed that ``\emph{Cars Are the Worst Product Category We Have Ever Reviewed for Privacy}''. This review presents an in-depth data practices analysis for 25 major automotive original equipment manufacturers (OEMs) worldwide, revealing the extensive range of data collected and shared by most automotive OEMs. The work carried out was manually done using privacy policies as the main data sources, and it also highlighted the ambiguity and misleading nature of the complex language used in privacy policies. Inspired by this public review, we decided to use the automotive industry as a case study to test the effectiveness and efficiency of \SysName\ in automating privacy policy analysis.

In this section, we first outline the methodology for selecting privacy policies as the dataset in Section~\ref{subsec:case_study_datasets}, followed by a description of the evaluation process in Section~\ref{subsec:case_study_evaluation}. Finally, the experimental results are presented and discussed in Section~\ref{subsec:case_study_results}.

\subsection{Datasets}
\label{subsec:case_study_datasets}

With the advancement of IoT and AI technologies, many modern vehicles offer connected vehicle services through mobile applications. These mobile apps offer various features to enhance driving assistance and user experience. Each mobile app should have a dedicated privacy policy to inform consumers of its data-handling practices. To systematically analyse the privacy policies governing these services, we implemented the following inclusion and exclusion criteria for selecting privacy policies:
\begin{enumerate}
\item The privacy policy must be published for a connected vehicle mobile app available in the EU or the UK. This selection criterion is motivated by the fact that legal frameworks and regulatory requirements governing data protection (e.g., the EU/UK GDPR) are closely aligned, leading to comparable structures and content and allowing systematic cross-policy analysis.
    
\item The privacy policy must be specifically and solely dedicated to the associated connected vehicle mobile app. If the identified policy for a connected vehicle mobile app was identical to its OEM's website general privacy policy, it was excluded. Since the general privacy policy focuses on the data handling practices of using the OEM's website rather than the connected vehicle mobile app.
    
\item The text of the privacy policy must be embedded within an HTML page rather than being provided as a linked document in formats such as PDF or Word. While LLMs can process other formats, the customised text-segmentation pipeline is optimised for HTML, which is also the dominant format for most of privacy policies.

\end{enumerate}

In addition to the above criteria, automotive OEMs and their associated apps with broken or inaccessible privacy policy links were excluded from the study. Using the Mozilla study~\cite{Caltrider-J2023} as an initial reference for selecting privacy policies, we further refined our dataset by applying the inclusion and exclusion criteria. The selected privacy policies\footnote{{The dataset can be accessed from~\url{https://osf.io/zgacu/overview?view_only=ee487642d88f4ce1a14473b8402d4762}}} are listed in Table~\ref{tab:privacy_policies}, where each privacy policy is saved as a single HTML file.

\begin{table}[!htb]
\centering
\caption{List of OEMs and their associated connected vehicle mobile apps}
\begin{tabular}{cc || cc}
\toprule
\textbf{OEM} & \textbf{Link} & \textbf{OEM} & \textbf{Link}\\
\midrule
Honda & \href{https://www.honda.co.uk/cars/owners/my-honda-plus/privacy-policy.html}{My Honda+}&
Lexus & \href{https://www.toyota-europe.com/legal/privacy-policy}{Lexus Link+}\\
Kia & \href{https://www.kia.com/uk/kia-connect-privacy-policy/}{Kia Connect} &
Nissan & \href{https://www.nissan.co.uk/gdpr.html}{NissanConnect Services}\\
Audi & \href{https://www.audi.com/en/privacy-policy.html}{Audi Privacy Policy} &
Vauxhall & \href{https://vx-mym.servicesgp.mpsa.com/webview/gdpr/privacy?culture=en-GB}{MyVauxhall}\\
Hyundai & \href{https://www.hyundai.com/eu/bluelink-privacy-notice}{Hyundai Bluelink Europe} &
Polestar & \href{https://www.polestar.com/global/legal/privacy/privacy-notice-polestar-app/}{Polestar}\\
Ford & \href{https://www.fordpass.com/content/ford_com/fp_app/en_gb/privacy.html}{FordPass}
& Renault & \href{https://www.renaultgroup.com/nos-engagements/le-groupe-renault-et-vos-donnees-personnelles/}{My Renault}\\
\bottomrule
\end{tabular}
\label{tab:privacy_policies}
\end{table}

\subsection{Evaluation}
\label{subsec:case_study_evaluation}

Due to the lack of ground truth datasets, we used manual validation to evaluate the effectiveness and correctness of \SysName. The first three co-authors of this study participated in the evaluation work as domain experts, where each independently evaluated the outputs generated by the proposed framework based on given text segments of a privacy policy. For each text segment, \SysName\ generates six outputs corresponding to the data type, data category, data flow, data consumer type, data processing purpose, and data processing method. In addition, the corresponding knowledge bases, as described in Section~\ref{subsubsec:pre-processor_DesignStepB_knowledge_bases} were shared with all evaluators as references to facilitate the evaluation. For each output, the evaluator was instructed to give a score using a 1-7 Likert scale (i.e., 1: Strongly disagree, 2: Disagree, 3: Somewhat disagree, 4: Neither agree nor disagree, 5: Somewhat agree, 6: Agree, 7: Strongly agree) to verify its relevance, correctness, and clarity using the given knowledge bases as references.

For the evaluation experiment, we randomly sampled approximately 40\% of the total outputs derived from the analysis of the My Honda+ app's privacy policy. This resulted in 150 output tuples, with each tuple containing six tasks, amounting to a total of 900 evaluation tasks per evaluator. The main reasons for choosing My Honda+ are as follows: 1) it produces the highest number of extracted data flows (see Section~\ref{subsec:case_study_results} for more details); and 2) the structure of the privacy policy text includes diverse formats, including tables, bullet points, and narrative paragraphs, providing a robust and varied evaluation set.

\begin{table}[!htb]
\centering
\caption{Evaluation scores by three evaluators (mean with standard deviation), 7-Likert scale}
\label{tab:evaluation_results}
\resizebox{\linewidth}{!}{ %
\begin{tabular}{lcccccc}
\toprule
\textbf{Evaluator} & \textbf{Data Type} & \textbf{Data Category} & \textbf{Data Flow} & \textbf{Data Consumer Type} & \textbf{Data Processing Purpose} & \textbf{Data Processing Method}\\
\midrule
1 & 6.74 (0.87) & 6.11 (1.35) & 6.27 (1.21) & 6.19 (1.39) & 6.62 (1.07) & 6.78 (0.93)\\
\midrule
2 & 6.97 (0.29) & 5.62 (1.95) & 6.52 (1.25) & 6.58 (1.20) & 6.28 (1.40) & 6.58 (1.10)\\
\midrule
3 & 6.95 (0.25) & 5.29 (2.06) & 6.02 (1.55) & 6.29 (1.58) & 4.60 (1.93) & 5.92 (1.48)\\
\bottomrule
\end{tabular}
}
\end{table}

Table~\ref{tab:evaluation_results} presents a summary of the evaluation scores across the six tasks assessed by the three evaluators. Overall, the results demonstrate that evaluators generally ``agree'' or ``strongly agree'' with the LLMs' outputs, with most average scores between 6 and 7. Specifically, the `Data Type' and `Data Flow' scores received consistently high ratings across all evaluators. The mean values ranged from 6.27 to 6.97, with relatively low standard deviations, indicating strong and consistent confidence in \SysName's ability to identify and classify these elements correctly. However, slight variations were observed in the evaluation of the `Data Category' and `Data Processing Purpose'. In particular, Evaluator 3 assigned a noticeably lower average score of 4.60 (SD = 1.93) for the purpose category. Similarly, the `Data category' task received the lower average score, particularly from Evaluators 2 and 3 (5.62 and 5.29, respectively).

The observed disparities could be caused by 1) \SysName's capability to classify data categories and data processing purposes may be less robust or more ambiguous in certain contexts, and 2) the definitions within the relevant knowledge bases used for these classifications may be inherently ambiguous, resulting in different interpretations between human evaluators and the LLMs. These highlight the need to improve \SysName's performance and the need to better define and construct knowledge bases. Nevertheless, the consistently high scores across most evaluation dimensions confirm the proposed framework's overall effectiveness in extracting structured and meaningful information from unstructured privacy policy texts.

\begin{table}[!htb]
\centering
\caption{Gwet's AC1 and percentage agreement using 7-Likert scale and transformed 3-category scale}
\label{tab:agreement}
\resizebox{\linewidth}{!}{ %
\begin{tabular}{lcccccc}
\toprule
\textbf{Metric} & \textbf{Data Type} & \textbf{Data Category} & \textbf{Data Flow} & \textbf{Data Consumer Type} & \textbf{Data Processing Purpose} & \textbf{Data Processing Method}\\
\midrule
\multicolumn{7}{l}{\emph{7-Likert scale}}\\
Gwet's AC1 & 0.85 & 0.37 & 0.51 & 0.56 & 0.21 & 0.56\\
Percent. Agreement & 0.89 & 0.43 & 0.59 & 0.66 & 0.33 & 0.66\\
\midrule
\multicolumn{7}{l}{\emph{Transformed 3-category scale}}\\
Gwet's AC1 & 0.94 & 0.56 & 0.82 & 0.83 & 0.48 & 0.79\\
Percent. Agreement & 0.96 & 0.59 & 0.86 & 0.86 & 0.53 & 0.83\\
\bottomrule
\end{tabular}
}
\end{table}

Moreover, it is essential to examine the inter-rater reliability metric to obtain more insights into the manual verification results. Since we observed that there could be high-agreement situations where traditional metrics such as Cohen's kappa coefficient and Intraclass Correlation Coefficient (ICC) might fail to reflect consensus among raters~\cite{Gwet-L2008, Feinstein-A1990}, we used both Gwet's AC1 and the percentage agreements in this study~\cite{Gwet-L2008, Hallgren-K2012}. Gwet's AC1 is a statistical measure of inter-rater reliability, and the percentage agreement measures the raw consensus among evaluators. For both metrics, the value ranges from 0 to 1, where 1 indicates perfect agreement.

Because the results were evaluated using a 7-point Likert scale, it was designed to capture subtle differences in agreements and disagreements. While this granularity is valuable for in-depth assessment, it naturally reduces the likelihood of reaching exact agreements among evaluators. Nevertheless, the agreement metrics based on the raw scores remain promising. As shown in Table~\ref{tab:agreement}, the overall agreement for the `Data Type' is particularly high, with Gwet’s AC1 at 0.85 and a corresponding percentage agreement of 0.889, indicating a strong consensus. However, the agreement was more moderate for other tasks. For `Data Flow', `Data Consumer Type', and `Data Processing Method', Gwet’s AC1 scores ranged between 0.5 and 0.56, with percentage agreement above 0.58, suggesting a reasonable level of alignment. The `Data Category' and `Data Processing Purpose' showed the lowest agreement, with Gwet’s AC1 at 0.37 and 0.21, respectively.

To better measure agreement at a broader level, we also applied a score transformation by collapsing the 7-point Likert scale into a 3-category scale: scores of 1–3 were mapped to 1 (disagreement), 4 to 2 (neutral), and 5–7 to 3 (agreement). This scoring scheme could reduce sensitivity to minor differences in evaluators' responses and produce a more robust measure of general agreement. As shown in Table~\ref{tab:agreement}, we observe an improvement in both metrics across all tasks. For instance, `Data Type' and `Data Flow' have Gwet's AC1 scores of 0.94 and 0.82, respectively, and corresponding percentage agreements above 0.86, indicating strong agreements on the general validity of \SysName's outputs. The Gwet's AC1 and percentage agreement for `Data Consumer Type' and `Data Processing Method' were approximately 0.8, suggesting strong agreements between evaluators. Although the scores for `Data Category' and `Data Processing Purpose' improved by approximately 20\%, the agreements were relatively low compared with other tasks. The results suggest that, while evaluators may differ on specific levels of agreement, there is a broader consensus on the effectiveness and accuracy of the proposed framework's capability in analysing complex privacy policy documents and extracting comprehensive data flows.

\subsection{Analysis of Results}
\label{subsec:case_study_results}

We aim to illustrate how interpretation and analysis can be conducted from different perspectives to generate insights of potential privacy concerns that might otherwise be hidden or neglected. For the convenience and clarity, we use OEMs' names instead of mobile app names throughout this section.

\subsubsection{Analysis of Data Flow Graphs}
\label{subsubsec:case_study_results_network_analysis}

\begin{table}[!htb]
\centering
\caption{Summary of data flow network statistics}
\label{tab:network_graph_stats}
\resizebox{\linewidth}{!}{ %
\begin{tabular}{lcccccccccc}
\toprule
& \textbf{Audi} & \textbf{Ford} & \textbf{Honda} & \textbf{Hyundai} & \textbf{Kia} & \textbf{Lexus} & \textbf{Nissan} & \textbf{Polestar} & \textbf{Renault} & \textbf{Vauxhall}\\
\midrule
Edges & 84 & 198 & 510 & 79 & 93 & 61 & 201 & 151 & 8 & 176\\
First-party nodes & 5 & 11 & 6 & 13 & 4 & 2 & 9 & 4 & 2 & 16\\
Third-party nodes & 22 & 33 & 51 & 19 & 16 & 17 & 42 & 14 & 0 & 23\\
User-party nodes & 2 & 3 & 3 & 3 & 2 & 1 & 3 & 3 & 1 & 1\\
Data types nodes & 22 & 73 & 177 & 14 & 32 & 15 & 64 & 64 & 4 & 35\\
\bottomrule
\end{tabular}
}
\end{table}

\begin{figure}[!htb]
\centering
\subfloat[]{\includegraphics[width=0.5\linewidth]{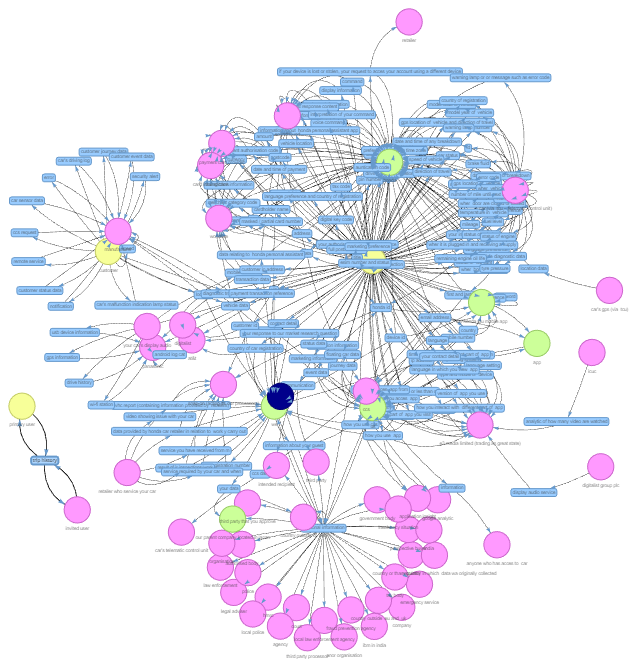}}%
\subfloat[]{\includegraphics[width=0.5\linewidth]{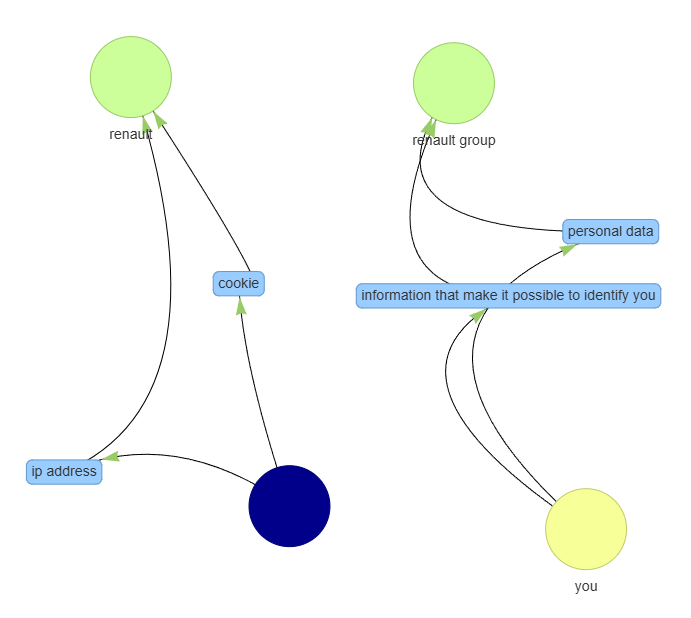}}%
\caption{Comparison between (a) a complex data flow network derived from My Honda+ app's privacy policy and (b) a simple data flow network derived from My Renault app's privacy policy}
\label{fig:graph_comparison}
\end{figure}

We represent data flows as graphs and conduct graph-based analysis. Table~\ref{tab:network_graph_stats} presents the descriptive statistics of the data flow graph networks extracted from the mobile apps' privacy policies of different OEMs. Honda has the most complex network, which contains 510 edges and more than 200 nodes. In contrast, Ford, Nissan, Polestar, and Vauxhall had moderately smaller networks, with the number of edges ranging from 170 to 270. Audi, Hyundai, Kia, and Lexus have relatively simpler network structures, whereas Renault has the smallest networks. Figure~\ref{fig:graph_comparison} presents a comparison between a complex network (Honda) and a simpler network (Renault) to illustrate the network structures derived from automated privacy policy analysis\footnote{All generated network graphs can be viewed from \url{https://osf.io/zgacu/overview?view_only=ee487642d88f4ce1a14473b8402d4762}}. In the visualisation, pink, yellow, and green nodes represent third-party, user-party, and first-party attributed nodes, respectively. For simplicity, these are referred to as third-party, user-party, and first-party nodes, respectively, in the rest of this paper. The nodes in the light blue boxes represent different types of data. Dark blue is used to annotate nodes with unknown classifications.

\begin{figure}[!htb]
\centering
\includegraphics[width=.8\linewidth]{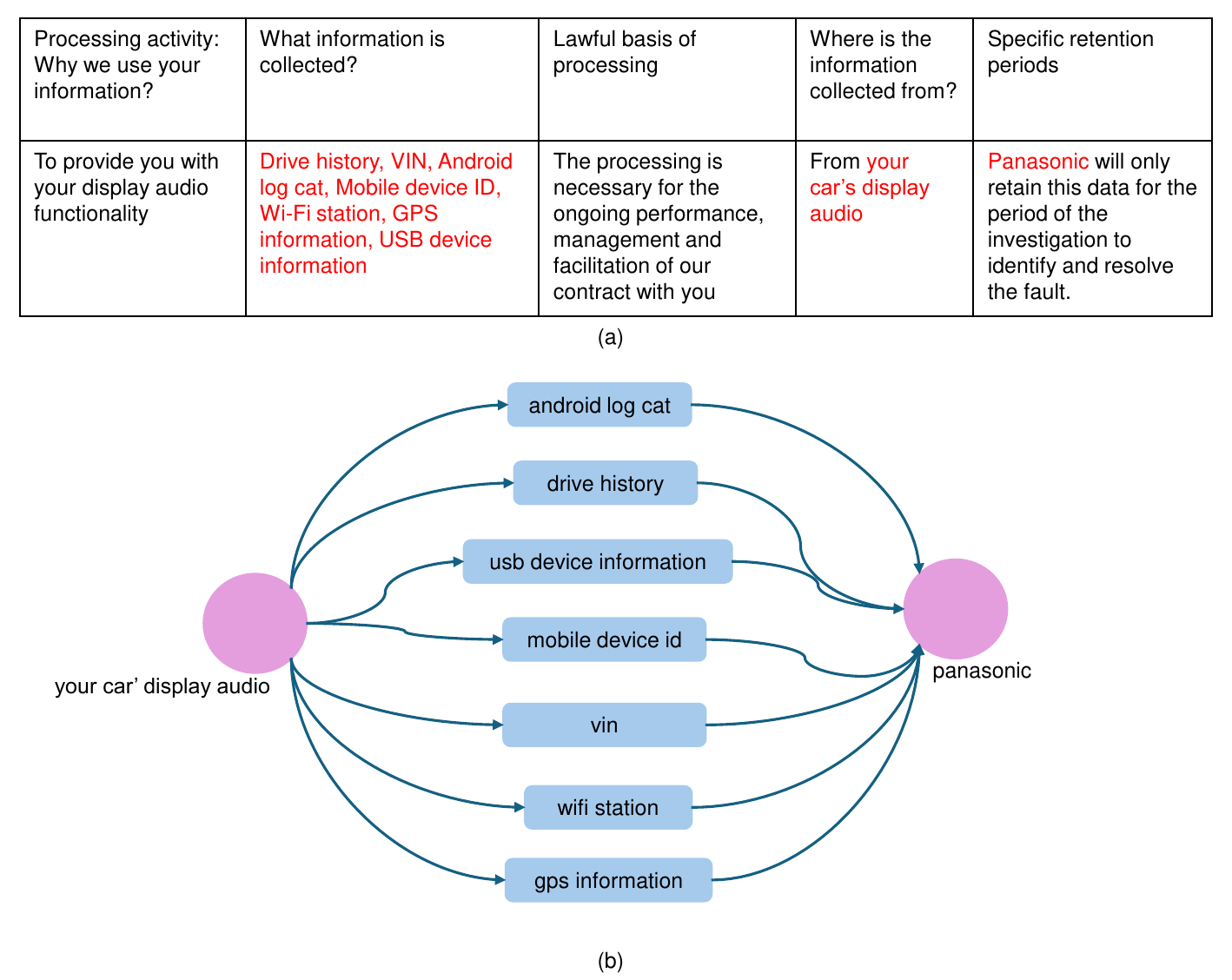}
\caption{Example of converting table content to data flows illustrated in part of a data flow network}
\label{fig:honda_panasonic}
\end{figure}

By taking a closer look at specific nodes as illustrated in Figure~\ref{fig:honda_panasonic}(b), the arrows reflect the directional nature of the network graph, indicating the direction of how data flows between entities. For instance, a third-party node `your car's display audio' would share an array of data, including `Android log cat', `drive history', `USB device information', `mobile device ID', `VIN', `WiFi station'/footnote{This is the exact wording used in the privacy policy. We understand that the actual meaning is WiFi station ID.}, and `GPS information' to a third-party node `Panasonic'. Additionally, Figure~\ref{fig:honda_panasonic}(a) shows a table of the raw text originally appearing in the privacy policy, demonstrating that our method can accurately extract granular information from HTML table content.

Moreover, to systematically compare networks and obtain a more comprehensive understanding, network metrics, including betweenness, closeness, and degree centrality, were computed, and the top ten nodes with the highest scores across all apps are listed in Table~\ref{tab:top_10_metrics}. To facilitate the analysis and visualisation, we use distinct colours to represent different categories of nodes. It is also worth noting that we chose not to merge or normalise nodes that share the same semantic meaning. For example, the nodes `We' and `Us' are semantically equivalent, as are `Personal data' and `Your data'. We retained them in their original wording to preserve fidelity to the source text and to minimise the risk of misclassification that may arise during such merging/normalisation process. To retain such raw form also allows us to directly loop up each node in the privacy policy text for cross-validation. Nevertheless, to address potential issues of semantic overlapping, we added additional attributes, including first-party, third-party, and user-party, to collection party nodes, as explained earlier in Section~\ref{subsubsec:post-processor_data_parser}.

\paragraph{Betweenness Centrality}
Betweenness centrality measures the extent to which a node lies on the shortest paths between other nodes and is often considered a ``bridge'' or intermediary in the network. As shown in Table~\ref{tab:top_10_metrics}, the top ten nodes with the highest betweenness centrality scores consistently include first-party nodes (e.g., OEMs names, `We', `Us'), user-party nodes (e.g., `You', `User') and data nodes for most apps. The only exceptions are Ford and Nissan, where Ford includes third-party service providers such as `Vodafone' and the `Roadside assistance provider', while Nissan has `Our partner' as a third party node in its top ten nodes. This implies the significant role of third-party service providers in Ford's and Nissan's data flow network.

Overall, the `Personal data'/`Personal information' node had the highest frequency, appearing in 9/10 apps' top ten lists. The `VIN' node also appeared in more than half of all apps. Both `personal data' and `VIN' act as critical bridges across different data flow networks. Moreover, there are some specific sensitive data types that are worth noticing. For instance, the finance data `Masked/partial card number' for Honda, biometric data `Voice recording' for Kia, personal identifiable data `Identity information' for Lexus, and health data `Medical personal data' for Nissan. The significance of these data in relation to betweenness centrality highlights the scale of sensitive data processing and their bridging roles in associated data flow networks.

\paragraph{Closeness Centrality}
Closeness centrality measures how close a node is to all other nodes in the network, highlighting its role in the information transmission of a network. By reviewing coloured stacked bars across all apps, Kia appears to have a distinct pattern, as most of its top ten nodes of closeness centrality are third-party nodes, where nodes, such as `Kia Connected GmbH' (associated with Germany),  `Cerence B.V.' (associated with the Netherlands), and `Recipient outside the EU/EEA' indicate the extensive cross-border information flow.

Hyundai and Lexus share similar patterns of distribution of node types. The top ten nodes of the Hyundai network graph includes nodes, such as `Hyundai Motor Company', `Hyundai entity located in the Republic of Korea', `Hyundai AutoEver Europe GmbH', `Hyundai AutoEver Corp.', and `Hyundai BluneLink Europe', that are affiliated with Hyundai Group. Similar to Hyundai, Lexus is part of the Toyota corporate Group, and we can see `Toyota' associated nodes such as `Toyota Financial Service', `Toyota Insurance Management', and `Toyota Financial Service', appear in the list of closeness centrality top ten nodes. These observations suggest that Hyundai and Lexus might have relatively better control over data flows within their networks if we assume that data processing practices could be more transparent within the same corporate group than those involving more third-party entities. 

It is worth noting that most of the apps' data flow network graphs are associated with third-party nodes. For instance, `Processor in third country', `Tracking service provider', `Web agency', `Hosting provider', and `IT service provider' are heavily involved in the data flow network of Audi. Similarly, `Analytic', `Business partner', and `Subcontractor' are key nodes in Polestar’s network. `Dealer', `Google', `Our partner', `RCI Financial Service Limited trading as Mobilize Financial Service' are essential third-party nodes in the data flow network of Nissan. This highlights the significant dependence on third-party entities to transmit information within these networks. However, the only exception is `Renault', which does not have a single third-party node appearing in the top ten list. This may indicate either a comparatively minimal reliance on third-party entities for service provision, or that the privacy policy is written in a way that omits or obscures such details.

\begin{table}[!htb]
\tiny
\centering
\caption{Top 10 betweenness, closeness, and degree of centrality network metrics}
\label{tab:top_10_metrics}
\begin{threeparttable}
\begin{tabularx}{\linewidth}{p{0.5cm} p{0.5cm} X p{2cm}}
\toprule
\textbf{OEM} & \textbf{Metric} & \textbf{Top 10 nodes in Descent order} & \textbf{Distribution of node types}\\
\midrule
Audi & BC & \sethlcolor{lightblue}\hl{Personal data}, \sethlcolor{lightgreen}\hl{Us}, \sethlcolor{lightblue}\hl{Data}, 
\sethlcolor{lightblue}\hl{IP address}, \sethlcolor{lightblue}\hl{Information}, \sethlcolor{lightblue}\hl{VIN}, \sethlcolor{lightblue}\hl{Your data}, \sethlcolor{lightblue}\hl{IP address}, \sethlcolor{lightblue}\hl{Information about cyber security vulnerabilities/incident or hacker attack}, \sethlcolor{lightblue}\hl{Contact detail}, \sethlcolor{lightblue}\hl{URL of visited websites} & \begin{tikzpicture}[baseline=(current bounding box.center)]\fill[lightblue] (0.000,0) rectangle (1.800,0.3);\fill[lightgreen] (1.800,0) rectangle (2.000,0.3);\draw (0,0) rectangle (2.0,0.3);\end{tikzpicture} \\
 & CC & \sethlcolor{lightblue}\hl{Log file}, \sethlcolor{lightgreen}\hl{Audi}, \sethlcolor{lightpink}\hl{Chatbot}, \sethlcolor{lightpink}\hl{Processor in third country},
 \sethlcolor{lightblue}\hl{Personal data}, \sethlcolor{lightpink}\hl{Processor}, \sethlcolor{lightpink}\hl{Tracking service provider}, \sethlcolor{lightpink}\hl{Web agency}, \sethlcolor{lightpink}\hl{Hosting provider}, \sethlcolor{lightpink}\hl{IT service provider} & \begin{tikzpicture}[baseline=(current bounding box.center)]\fill[lightblue] (0.000,0) rectangle (0.400,0.3);\fill[lightgreen] (0.400,0) rectangle (0.600,0.3);\fill[lightpink] (0.600,0) rectangle (2.000,0.3);\draw (0,0) rectangle (2.0,0.3);\end{tikzpicture} \\
 & DC & \sethlcolor{lightblue}\hl{Personal data}, \sethlcolor{lightpink}\hl{Internet browser}, \sethlcolor{lightblue}\hl{Log file}, \sethlcolor{lightblue}\hl{Data}, 
 \sethlcolor{lightblue}\hl{IP address}, \sethlcolor{lightyellow}\hl{User}, \sethlcolor{lightgreen}\hl{Audi}, \sethlcolor{lightpink}\hl{Chatbot}, \sethlcolor{lightgreen}\hl{Us}, \sethlcolor{lightblue}\hl{Your data} & \begin{tikzpicture}[baseline=(current bounding box.center)]
  \fill[lightblue] (0,0) rectangle (1.0,0.3);
  \fill[lightpink] (1.0,0) rectangle (1.4,0.3);
  \fill[lightyellow] (1.4,0) rectangle (1.6,0.3);
  \fill[lightgreen] (1.6,0) rectangle (2.0,0.3);
  \draw (0,0) rectangle (2.0,0.3);
\end{tikzpicture}\\
\midrule
Ford & BC & \sethlcolor{lightblue}\hl{VIN}, \sethlcolor{lightyellow}\hl{User}, \sethlcolor{lightblue}\hl{Data use threshold}, \sethlcolor{lightpink}\hl{Vodafone Global Enterprise Limited (Vodafone)}, \sethlcolor{lightgreen}\hl{Us},\sethlcolor{lightgreen}\hl{Ford Smart Mobility UK Limited (FSM)}, \sethlcolor{lightblue}\hl{Information that you have purchased the subscription}, \sethlcolor{lightpink}\hl{Roadside assistance provider (RSA)}, \sethlcolor{lightgreen}\hl{We}, \sethlcolor{lightblue}\hl{Vehicle location} & \begin{tikzpicture}[baseline=(current bounding box.center)]\fill[lightblue] (0.000,0) rectangle (0.889,0.3);\fill[lightgreen] (0.889,0) rectangle (1.333,0.3);\fill[lightpink] (1.333,0) rectangle (1.778,0.3);\fill[lightyellow] (1.778,0) rectangle (2.000,0.3);\draw (0,0) rectangle (2.0,0.3);\end{tikzpicture}\\
 & CC & \sethlcolor{lightgreen}\hl{Ford}, \sethlcolor{lightgreen}\hl{We}, \sethlcolor{lightgreen}\hl{Ford Smart Mobility UK Limited (FSM)}, \sethlcolor{lightpink}\hl{Google}, \sethlcolor{lightblue}\hl{Vehicle data}, \sethlcolor{lightgreen}\hl{Us}, \sethlcolor{lightblue}\hl{VIN}, \sethlcolor{lightpink}\hl{Our authorized dealer}, \sethlcolor{lightpink}\hl{Company or organisation}, \sethlcolor{lightblue}\hl{Some data} & \begin{tikzpicture}[baseline=(current bounding box.center)]\fill[lightblue] (0.000,0) rectangle (0.500,0.3);\fill[lightgreen] (0.500,0) rectangle (1.500,0.3);\fill[lightpink] (1.500,0) rectangle (2.000,0.3);\draw (0,0) rectangle (2.0,0.3);\end{tikzpicture}\\
 & DC & \sethlcolor{lightyellow}\hl{You}, \sethlcolor{lightgreen}\hl{Ford}, \sethlcolor{lightgreen}\hl{We}, \sethlcolor{lightyellow}\hl{User}, \sethlcolor{lightblue}\hl{Vehicle location}, \sethlcolor{lightgreen}\hl{Us}, \sethlcolor{lightgreen}\hl{Ford Smart Mobility UK Limited (FSM)}, \sethlcolor{lightpink}\hl{Google}, \sethlcolor{lightblue}\hl{Speed},  \sethlcolor{lightblue}\hl{Vehicle information} & \begin{tikzpicture}[baseline=(current bounding box.center)]
  \fill[lightblue] (0,0) rectangle (0.6,0.3);
  \fill[lightgreen] (0.6,0) rectangle (1.4,0.3);
  \fill[lightpink] (1.4,0) rectangle (1.6,0.3);
  \fill[lightyellow] (1.6,0) rectangle (2.0,0.3);
  \draw (0,0) rectangle (2.0,0.3);
\end{tikzpicture}\\
\midrule
Honda & BC & \sethlcolor{lightgreen}\hl{Honda}, \sethlcolor{lightyellow}\hl{You}, \sethlcolor{lightblue}\hl{Authentication code}, \sethlcolor{lightblue}\hl{Personal information}, \sethlcolor{lightgreen}\hl{We}, \sethlcolor{lightgreen}\hl{Us}, \sethlcolor{lightblue}\hl{VIN}, \sethlcolor{lightblue}\hl{Honda ID}, \sethlcolor{lightblue}\hl{Data}, \sethlcolor{lightblue}\hl{Masked/partial card number} & \begin{tikzpicture}[baseline=(current bounding box.center)]\fill[lightblue] (0.000,0) rectangle (1.111,0.3);\fill[lightgreen] (1.111,0) rectangle (1.778,0.3);\fill[lightpink] (1.778,0) rectangle (1.778,0.3);\fill[lightyellow] (1.778,0) rectangle (2.000,0.3);\fill[lightpurple] (2.000,0) rectangle (2.000,0.3);\draw (0,0) rectangle (2.0,0.3);\end{tikzpicture}\\
 & CC & \sethlcolor{lightgreen}\hl{Honda}, \sethlcolor{lightblue}\hl{Data}, \sethlcolor{lightpink}\hl{SoundHound}, \sethlcolor{lightblue}\hl{Authentication code}, \sethlcolor{lightblue}\hl{Personal information}, \sethlcolor{lightblue}\hl{Honda ID}, \sethlcolor{lightblue}\hl{VIN}, \sethlcolor{lightblue}\hl{Customer ID}, \sethlcolor{lightgreen}\hl{Us}, \sethlcolor{lightgreen}\hl{We} & \begin{tikzpicture}[baseline=(current bounding box.center)]\fill[lightblue] (0.000,0) rectangle (1.111,0.3);\fill[lightgreen] (1.111,0) rectangle (1.778,0.3);\fill[lightpink] (1.778,0) rectangle (2.000,0.3);\draw (0,0) rectangle (2.0,0.3);\end{tikzpicture}\\
 & DC & \sethlcolor{lightyellow}\hl{You}, \sethlcolor{lightgreen}\hl{Honda}, \sethlcolor{lightpink}\hl{Car (via TCU - Telematic Control Unit)}, \sethlcolor{lightblue}\hl{Personal information}, \sethlcolor{lightgreen}\hl{Us}, \sethlcolor{lightgreen}\hl{We}, \sethlcolor{lightpink}\hl{E3 Media Limited}, \sethlcolor{lightpink}\hl{Worldline}, \sethlcolor{lightyellow}\hl{Customer}, \sethlcolor{lightpink}\hl{Manufacturer} & \begin{tikzpicture}[baseline=(current bounding box.center)]\fill[lightblue] (0.000,0) rectangle (0.250,0.3);\fill[lightgreen] (0.250,0) rectangle (0.750,0.3);\fill[lightpink] (0.750,0) rectangle (1.500,0.3);\fill[lightyellow] (1.500,0) rectangle (2.000,0.3);\draw (0,0) rectangle (2.0,0.3);\end{tikzpicture}\\
\midrule
Hyundai & BC & \sethlcolor{lightblue}\hl{Personal data}, \sethlcolor{lightblue}\hl{Your personal data}, \sethlcolor{lightgreen}\hl{We}, \sethlcolor{lightgreen}\hl{Hyundai Motor Company}, \sethlcolor{lightgreen}\hl{Hyundai}, \sethlcolor{lightblue}\hl{Your data}, \sethlcolor{lightblue}\hl{Certain personal data}, \sethlcolor{lightblue}\hl{VIN}, \sethlcolor{lightblue}\hl{Security event-related data}, \sethlcolor{lightblue}\hl{Timestamp of the generated security event} & \begin{tikzpicture}[baseline=(current bounding box.center)]\fill[lightblue] (0.000,0) rectangle (1.333,0.3);\fill[lightgreen] (1.333,0) rectangle (2.000,0.3);\fill[lightpink] (2.000,0) rectangle (2.000,0.3);\fill[lightyellow] (2.000,0) rectangle (2.000,0.3);\draw (0,0) rectangle (2.0,0.3);\end{tikzpicture}\\
 & CC & \sethlcolor{lightpink}\hl{Data processor}, \sethlcolor{lightgreen}\hl{Hyundai Motor Company}, \sethlcolor{lightblue}\hl{Your personal data}, \sethlcolor{lightgreen}\hl{Hyundai entity located in the Republic of Korea}, \sethlcolor{lightpink}\hl{Recipient of your personal data}, \sethlcolor{lightpink}\hl{Cerence sub-processor}, \sethlcolor{lightgreen}\hl{Hyundai AutoEver Europe GmbH}, \sethlcolor{lightgreen}\hl{Hyundai AutoEver Corp.}, \sethlcolor{lightpink}\hl{Republic of Korea}, \sethlcolor{lightgreen}\hl{Hyundai BlueLink Europe} & \begin{tikzpicture}[baseline=(current bounding box.center)]\fill[lightblue] (0.000,0) rectangle (0.200,0.3);\fill[lightgreen] (0.200,0) rectangle (1.200,0.3);\fill[lightpink] (1.200,0) rectangle (2.000,0.3);\draw (0,0) rectangle (2.0,0.3);\end{tikzpicture}\\
 & DC & \sethlcolor{lightblue}\hl{Personal data}, \sethlcolor{lightyellow}\hl{You}, \sethlcolor{lightblue}\hl{Your personal data}, \sethlcolor{lightgreen}\hl{We}, \sethlcolor{lightblue}\hl{Certain personal data}, \sethlcolor{lightpink}\hl{Data processor}, \sethlcolor{lightgreen}\hl{Hyundai Motor Company}, \sethlcolor{lightgreen}\hl{Hyundai}, \sethlcolor{lightblue}\hl{VIN}, \sethlcolor{lightblue}\hl{Security event-related data} & \begin{tikzpicture}[baseline=(current bounding box.center)]\fill[lightblue] (0.000,0) rectangle (1.111,0.3);\fill[lightgreen] (1.111,0) rectangle (1.556,0.3);\fill[lightpink] (1.556,0) rectangle (1.778,0.3);\fill[lightyellow] (1.778,0) rectangle (2.000,0.3);\draw (0,0) rectangle (2.0,0.3);\end{tikzpicture}\\
\midrule
Kia & BC & \sethlcolor{lightblue}\hl{Personal data}, \sethlcolor{lightgreen}\hl{Kia}, \sethlcolor{lightblue}\hl{Your personal data}, \sethlcolor{lightblue}\hl{Log-in data}, \sethlcolor{lightblue}\hl{Location data (GPS)}, \sethlcolor{lightblue}\hl{Personal data relating to the contractual relationship}, \sethlcolor{lightblue}\hl{Communication}, \sethlcolor{lightblue}\hl{Commercial letter}, \sethlcolor{lightblue}\hl{Contact detail}, \sethlcolor{lightblue}\hl{Voice recording} & \begin{tikzpicture}[baseline=(current bounding box.center)]\fill[lightblue] (0.000,0) rectangle (1.778,0.3);\fill[lightgreen] (1.778,0) rectangle (2.000,0.3);\fill[lightpink] (2.000,0) rectangle (2.000,0.3);\fill[lightyellow] (2.000,0) rectangle (2.000,0.3);\draw (0,0) rectangle (2.0,0.3);\end{tikzpicture}\\
 & CC & \sethlcolor{lightgreen}\hl{Kia Connected GmbH}, \sethlcolor{lightpink}\hl{Cerence B.V.}, \sethlcolor{lightpink}\hl{Autorité de Protection de Donnée}, \sethlcolor{lightpink}\hl{Gegevensbeschermingsautoriteit}, \sethlcolor{lightpink}\hl{Government authority}, \sethlcolor{lightpink}\hl{Court}, \sethlcolor{lightpink}\hl{External advisor}, \sethlcolor{lightpink}\hl{Recipient outside the EU/EEA, Similar third-party that are public body}, \sethlcolor{lightpink}\hl{Service provider} & \begin{tikzpicture}[baseline=(current bounding box.center)]\fill[lightblue] (0.000,0) rectangle (0.000,0.3);\fill[lightgreen] (0.000,0) rectangle (0.222,0.3);\fill[lightpink] (0.222,0) rectangle (2.000,0.3);\draw (0,0) rectangle (2.0,0.3);\end{tikzpicture}\\
 & DC & \sethlcolor{lightyellow}\hl{User}, \sethlcolor{lightgreen}\hl{Kia Connected GmbH}, \sethlcolor{lightblue}\hl{Personal data}, \sethlcolor{lightgreen}\hl{Kia}, \sethlcolor{lightblue}\hl{Log-in data}, \sethlcolor{lightblue}\hl{Location data (GPS)}, \sethlcolor{lightpink}\hl{Cerence B.V.}, \sethlcolor{lightpink}\hl{Autorité de Protection de Donnée}, \sethlcolor{lightpink}\hl{Gegevensbeschermingsautoriteit}, \sethlcolor{lightblue}\hl{Voice recording} & \begin{tikzpicture}[baseline=(current bounding box.center)]\fill[lightblue] (0.000,0) rectangle (0.800,0.3);\fill[lightgreen] (0.800,0) rectangle (1.200,0.3);\fill[lightpink] (1.200,0) rectangle (1.800,0.3);\fill[lightyellow] (1.800,0) rectangle (2.000,0.3);\draw (0,0) rectangle (2.0,0.3);\end{tikzpicture}\\
 \midrule
Lexus & BC & \sethlcolor{lightblue}\hl{Personal data}, \sethlcolor{lightblue}\hl{Postcode}, \sethlcolor{lightblue}\hl{Address}, \sethlcolor{lightblue}\hl{Personal information}, \sethlcolor{lightblue}\hl{Name}, \sethlcolor{lightblue}\hl{Telephone number}, \sethlcolor{lightblue}\hl{Email address}, \sethlcolor{lightblue}\hl{VIN}, \sethlcolor{lightblue}\hl{Geo location}, \sethlcolor{lightblue}\hl{Identity information} & \begin{tikzpicture}[baseline=(current bounding box.center)]\fill[lightblue] (0.000,0) rectangle (2.000,0.3);\draw (0,0) rectangle (2.0,0.3);\end{tikzpicture}\\
 & CC & \sethlcolor{lightgreen}\hl{Us}, \sethlcolor{lightgreen}\hl{Toyota}, \sethlcolor{lightpink}\hl{Authorised staff member}, \sethlcolor{lightpink}\hl{Affiliate and subsidiary company}, \sethlcolor{lightgreen}\hl{Toyota Financial Service}, \sethlcolor{lightgreen}\hl{Toyota Insurance Management}, \sethlcolor{lightgreen}\hl{Toyota Insurance Manager}, \sethlcolor{lightpink}\hl{Authorised retailer}, \sethlcolor{lightpink}\hl{Authorised repairer}, \sethlcolor{lightblue}\hl{Personal data} & \begin{tikzpicture}[baseline=(current bounding box.center)]\fill[lightblue] (0.000,0) rectangle (0.222,0.3);\fill[lightgreen] (0.222,0) rectangle (1.111,0.3);\fill[lightpink] (1.111,0) rectangle (2.000,0.3);\draw (0,0) rectangle (2.0,0.3);\end{tikzpicture}\\
 & DC & \sethlcolor{lightblue}\hl{Personal data}, \sethlcolor{lightyellow}\hl{You}, \sethlcolor{lightgreen}\hl{Us}, \sethlcolor{lightblue}\hl{Postcode}, \sethlcolor{lightblue}\hl{Address}, \sethlcolor{lightgreen}\hl{Toyota Financial Service}, \sethlcolor{lightblue}\hl{Personal information}, \sethlcolor{lightpink}\hl{Authorised staff member}, \sethlcolor{lightpink}\hl{Affiliate and subsidiary company}, \sethlcolor{lightpink}\hl{Member of our authorised retailer and authorised repairer network} & \begin{tikzpicture}[baseline=(current bounding box.center)]\fill[lightblue] (0.000,0) rectangle (0.800,0.3);\fill[lightgreen] (0.800,0) rectangle (1.200,0.3);\fill[lightpink] (1.200,0) rectangle (1.800,0.3);\fill[lightyellow] (1.800,0) rectangle (2.000,0.3);\draw (0,0) rectangle (2.0,0.3);\end{tikzpicture}\\
\midrule
Nissan & BC & \sethlcolor{lightblue}\hl{Personal data}, \sethlcolor{lightgreen}\hl{Us}, \sethlcolor{lightgreen}\hl{We}, \sethlcolor{lightpink}\hl{Our partner}, \sethlcolor{lightblue}\hl{Your Facebook website browsing data}, \sethlcolor{lightblue}\hl{Anonymised statistical data}, \sethlcolor{lightblue}\hl{Medical personal data}, \sethlcolor{lightblue}\hl{Data provided directly}, \sethlcolor{lightblue}\hl{Data relating to browsing our website}, \sethlcolor{lightblue}\hl{Data relating to use of our mobile application},  & \begin{tikzpicture}[baseline=(current bounding box.center)]
  \fill[lightblue] (0,0) rectangle (1.4,0.3);
  \fill[lightgreen] (1.4,0) rectangle (1.8,0.3);
  \fill[lightpink] (1.8,0) rectangle (2.0,0.3);
  \draw (0,0) rectangle (2.0,0.3);
\end{tikzpicture}\\
 & CC & \sethlcolor{lightgreen}\hl{Nissan}, \sethlcolor{lightpink}\hl{Any authorised dealer or repairer}, \sethlcolor{lightpink}\hl{Our financial partner}, \sethlcolor{lightgreen}\hl{Us}, \sethlcolor{lightblue}\hl{Personal data}, \sethlcolor{lightpink}\hl{Dealer}, \sethlcolor{lightgreen}\hl{We}, \sethlcolor{lightgreen}\hl{Nissan Automotive Europe S.A.S}, \sethlcolor{lightgreen}\hl{Nissan Motor (GB) Limited}, \sethlcolor{lightpink}\hl{Nissan Automotive S.A.S} & \begin{tikzpicture}[baseline=(current bounding box.center)]
  \fill[lightblue] (0,0) rectangle (0.2,0.3);
  \fill[lightgreen] (0.2,0) rectangle (1.2,0.3);
  \fill[lightpink] (1.2,0) rectangle (2.0,0.3);
  \draw (0,0) rectangle (2.0,0.3);
\end{tikzpicture}\\
 & DC & \sethlcolor{lightblue}\hl{Personal data}, \sethlcolor{lightyellow}\hl{You}, Unknown, \sethlcolor{lightgreen}\hl{Nissan}, \sethlcolor{lightpink}\hl{Any authorised dealer or repairer}, \sethlcolor{lightpink}\hl{Our financing partner}, \sethlcolor{lightgreen}\hl{Us}, \sethlcolor{lightpink}\hl{Our partner}, \sethlcolor{lightgreen}\hl{Nissan Automotive Europe S.A.S}, \sethlcolor{lightgreen}\hl{We}, \sethlcolor{lightpink}\hl{Dealer} & \begin{tikzpicture}[baseline=(current bounding box.center)]
  \fill[lightblue] (0,0) rectangle (0.2,0.3);
  \fill[lightgreen] (0.2,0) rectangle (1.0,0.3);
  \fill[lightpink] (1.0,0) rectangle (1.8,0.3);
  \fill[lightyellow] (1.8,0) rectangle (2.0,0.3);
  \draw (0,0) rectangle (2.0,0.3);
\end{tikzpicture}\\
\midrule

Polestar & BC & \sethlcolor{lightgreen}\hl{Polestar App}, \sethlcolor{lightgreen}\hl{We}, 
\sethlcolor{lightblue}\hl{Personal data}, \sethlcolor{lightblue}\hl{Phone number}, \sethlcolor{lightblue}\hl{Email address},
\sethlcolor{lightblue}\hl{Your information}, \sethlcolor{lightblue}\hl{VIN}, \sethlcolor{lightblue}\hl{Location}, \sethlcolor{lightblue}\hl{Contact information}, \sethlcolor{lightblue}\hl{Home address} & \begin{tikzpicture}[baseline=(current bounding box.center)]\fill[lightblue] (0.000,0) rectangle (1.600,0.3);\fill[lightgreen] (1.600,0) rectangle (2.000,0.3);\draw (0,0) rectangle (2.0,0.3);\end{tikzpicture} \\
& CC & \sethlcolor{lightpink}\hl{Analytic}, \sethlcolor{lightgreen}\hl{We}, \sethlcolor{lightpink}\hl{Customer support}, \sethlcolor{lightpink}\hl{Vehicl control}, \sethlcolor{lightblue}\hl{Personal data}, 
\sethlcolor{lightblue}\hl{Your information}, \sethlcolor{lightgreen}\hl{Polestar app}, 
\sethlcolor{lightgreen}\hl{Polestar},
\sethlcolor{lightpink}\hl{Business partner}, \sethlcolor{lightpink}\hl{Subcontractor} & \begin{tikzpicture}[baseline=(current bounding box.center)] \fill[lightblue] (0,0) rectangle (0.4,0.3); \fill[lightgreen] (0.4,0) rectangle (1,0.3); \fill[lightpink] (1,0) rectangle (2.0,0.3); \draw (0,0) rectangle (2.0,0.3);\end{tikzpicture}\\
& DC & \sethlcolor{lightpurple}\hl{Unkonwn}, \sethlcolor{lightpink}\hl{Analytic}, \sethlcolor{lightgreen}\hl{Polestar app}, \sethlcolor{lightgreen}\hl{We}, \sethlcolor{lightyellow}\hl{Customer}, \sethlcolor{lightblue}\hl{Personal data},\sethlcolor{lightpink}\hl{Customer support}, \sethlcolor{lightpink}\hl{Vehicl control}, \sethlcolor{lightblue}\hl{Phone number}, \sethlcolor{lightblue}\hl{Email address}, & \begin{tikzpicture}[baseline=(current bounding box.center)] \fill[lightblue] (0,0) rectangle (0.66,0.3); \fill[lightgreen] (0.66,0) rectangle (1,0.3); \fill[lightpink] (1,0) rectangle (1.66,0.3); \fill[lightyellow] (1.66,0) rectangle (1.83,0.3); \fill[lightpurple] (1.83,0) rectangle (2.0,0.3); \draw (0,0) rectangle (2.0,0.3);\end{tikzpicture}\\

\midrule
Renault & BC & \sethlcolor{lightblue}\hl{Personal data}, \sethlcolor{lightblue}\hl{Information that makes it possible to identify you}, \sethlcolor{lightblue}\hl{IP address}, \sethlcolor{lightblue}\hl{Cookie}, \sethlcolor{lightyellow}\hl{You}, \sethlcolor{lightgreen}\hl{Renault Group}, \sethlcolor{lightgreen}\hl{Renault} & \begin{tikzpicture}[baseline=(current bounding box.center)]\fill[lightblue] (0.000,0) rectangle (1.143,0.3);\fill[lightgreen] (1.143,0) rectangle (1.714,0.3);\fill[lightyellow] (1.714,0) rectangle (2.000,0.3);\draw (0,0) rectangle (2.0,0.3);\end{tikzpicture}\\
 & CC & \sethlcolor{lightgreen}\hl{Renault Group}, \sethlcolor{lightgreen}\hl{Renault}, \sethlcolor{lightblue}\hl{Personal data}, \sethlcolor{lightblue}\hl{Information that makes it possible to identify you}, \sethlcolor{lightblue}\hl{IP address}, \sethlcolor{lightblue}\hl{Cookie}, \sethlcolor{lightyellow}\hl{You}, \sethlcolor{lightpurple}\hl{Unknown} & \begin{tikzpicture}[baseline=(current bounding box.center)]
  \fill[lightgreen] (0,0) rectangle (0.5,0.3);
  \fill[lightblue] (0.5,0) rectangle (1.5,0.3);
  \fill[lightyellow] (1.5,0) rectangle (1.75,0.3);
  \fill[lightpurple] (1.75,0) rectangle (2.0,0.3);
  \draw (0,0) rectangle (2.0,0.3);
\end{tikzpicture}\\
 & DC & \sethlcolor{lightyellow}\hl{You}, \sethlcolor{lightblue}\hl{Personal data}, \sethlcolor{lightgreen}\hl{Renault Group}, \sethlcolor{lightblue}\hl{Information that makes it possible to identify you}, \sethlcolor{lightpurple}\hl{Unknown}, \sethlcolor{lightblue}\hl{IP address}, \sethlcolor{lightgreen}\hl{Renault}, \sethlcolor{lightblue}\hl{Cookie} & \begin{tikzpicture}[baseline=(current bounding box.center)]\fill[lightblue] (0.000,0) rectangle (0.857,0.3);\fill[lightgreen] (0.857,0) rectangle (1.429,0.3);\fill[lightyellow] (1.429,0) rectangle (1.714,0.3);\fill[lightpurple] (1.714,0) rectangle (2.000,0.3);\draw (0,0) rectangle (2.0,0.3);\end{tikzpicture}\\
\midrule
Vauxhall & BC & \sethlcolor{lightpink}\hl{Data processor}, \sethlcolor{lightblue}\hl{Personal data}, \sethlcolor{lightgreen}\hl{We}, \sethlcolor{lightblue}\hl{Contact detail}, \sethlcolor{lightgreen}\hl{Our network}, \sethlcolor{lightblue}\hl{Your data}, \sethlcolor{lightblue}\hl{Aggregated information}, \sethlcolor{lightblue}\hl{Vehicle data}, \sethlcolor{lightblue}\hl{Data inferred by our activity}, \sethlcolor{lightblue}\hl{Data collected by the browser} &  \begin{tikzpicture}[baseline=(current bounding box.center)]\fill[lightblue] (0.000,0) rectangle (1.333,0.3);\fill[lightgreen] (1.333,0) rectangle (1.778,0.3);\fill[lightpink] (1.778,0) rectangle (2.000,0.3);\draw (0,0) rectangle (2.0,0.3);\end{tikzpicture}\\
 & CC & \sethlcolor{lightgreen}\hl{Us}, \sethlcolor{lightpink}\hl{Stellantis Europe}, \sethlcolor{lightpink}\hl{Partner}, \sethlcolor{lightgreen}\hl{We}, \sethlcolor{lightgreen}\hl{Car manufacturer}, \sethlcolor{lightpink}\hl{Third selected partner}, \sethlcolor{lightpink}\hl{Social media platform}, \sethlcolor{lightpink}\hl{Programmatic advertising platform}, \sethlcolor{lightblue}\hl{Contact detail}, \sethlcolor{lightgreen}\hl{Our website and application} & \begin{tikzpicture}[baseline=(current bounding box.center)]\fill[lightblue] (0.000,0) rectangle (0.222,0.3);\fill[lightgreen] (0.222,0) rectangle (0.889,0.3);\fill[lightpink] (0.889,0) rectangle (2.000,0.3);\draw (0,0) rectangle (2.0,0.3);\end{tikzpicture}\\
 & DC & \sethlcolor{lightblue}\hl{Data}, \sethlcolor{lightyellow}\hl{You}, \sethlcolor{lightgreen}\hl{Us}, \sethlcolor{lightblue}\hl{Personal data}, \sethlcolor{lightgreen}\hl{We}, \sethlcolor{lightblue}\hl{Contact detail}, \sethlcolor{lightpink}\hl{Partner}, \sethlcolor{lightgreen}\hl{Your device}, \sethlcolor{lightpink}\hl{Stellantis Europe}, \sethlcolor{lightpink}\hl{Social media platform} & \begin{tikzpicture}[baseline=(current bounding box.center)]\fill[lightblue] (0.000,0) rectangle (0.600,0.3);\fill[lightgreen] (0.600,0) rectangle (1.200,0.3);\fill[lightpink] (1.200,0) rectangle (1.800,0.3);\fill[lightyellow] (1.800,0) rectangle (2.000,0.3);\draw (0,0) rectangle (2.0,0.3);\end{tikzpicture}\\
\bottomrule
\end{tabularx}
\begin{tablenotes}
\tiny
\item BC: Betweenness Centrality, CC: Closeness Centrality, DC: Degree Centrality
\item Colour scheme: \sethlcolor{lightblue}\hl{Data type nodes}, \sethlcolor{lightgreen}\hl{first-party nodes}, \sethlcolor{lightpink}\hl{third-party nodes}, \sethlcolor{lightyellow}\hl{User-party nodes}, \sethlcolor{lightpurple}\hl{Unknown nodes}
\end{tablenotes}
\end{threeparttable}
\end{table}

\paragraph{Degree Centrality}
Degree centrality reveals centralised hubs/nodes of the network by computing the number of direct connections within a network. To this end, a node with the highest degree of centrality indicates its directional connection to the most nodes, signifying its central role within a network. By inspecting the top ten nodes for all apps, the top three nodes typically included a first-party node (e.g., `We', `Us', or the OEM's name), a user-party node (e.g., `You', `User', `Customer'), and a data-type node (e.g., `Personal data', `Data'). Different from the observations for betweenness centrality and closeness centrality, the top ten nodes of degree centrality across all apps include all first-party, third-party, user-party, and data type nodes, suggesting a high level of interconnectivity across all node types within the data flow network graph.  

It is worth noting that `Unknown' nodes appear among the top ten nodes for both Polestar and Renault across different metrics. Manual examination of their privacy policies helped explain such nodes. For instance, for Renault's data flows `Unknown $\rightarrow$ Cookies $\rightarrow$ Renault' and `Unknown $\rightarrow$ IP address $\rightarrow$ Renault', the privacy policy states: ``\emph{For information relating to personal data that we automatically collect, such as IP address and cookies, a cookie policy is also available on each of Renault's websites or mobile applications.}'' While \SysName\ correctly identified `Cookies' and `IP address' as categories of personal data, it did not assign a specific data sender. This is consistent with the policy text, which does not explicitly state the sender of the data. One could reasonably infer that the sender is the user of the app or device, since IP addresses and cookies originate from the user's machine. However, because the input segment of the policy text does not provide explicit attribution, we adopt a conservative stance: \SysName’s classification of the sender as `Unknown' reflects the absence of clear textual evidence. Similar patterns were observed in some of the Polestar results. We acknowledge that with carefully tailored prompt design, LLMs could potentially infer such implicit roles more reliably, although this requires balancing inference with staying true to the text. In addition, we suggest that a more refined text‑segmentation strategy that incorporates more cross‑paragraph context may further improve the LLM’s ability. However, compared with feeding shorter text segments to LLMs, we also noticed that supplying overly lengthy text is not ideal for extracting fine-grained data flows, as it can incorrectly associate unrelated data senders/data receivers with data types. Further work is therefore needed to explore more dynamic and context‑aware segmentation strategies that balance contextual completeness with the precision required for detailed data‑flow extraction.

Moreover, as identified in Table~\ref{tab:network_graph_stats}, Renault has only three and four data type nodes, respectively. This is abnormal compared to other apps. We manually examined its privacy policy. For Renault, it applies a generic approach to describe its data handling practices throughout the privacy policy, where only generic terms such as `personal data' or `information' are used. This raises concerns about the transparency of its privacy policies, suggesting that the privacy policies may be ambiguous or lack clarity in explicitly explaining data-handling practices.

\paragraph{Summary}

Overall, graph analysis based on the centrality of betweenness, closeness, and degrees can enhance our understanding of data collection and sharing practices stated in privacy policies. Here, we summarise the insights obtained from such network analyses as follows:
\begin{itemize}
\item \textbf{Bridging role of data types in the data flow network}: `Personal data' and `VIN' (appearing in more than half of apps), and specific sensitive data (e.g., Honda's `masked card numbers', Kia's `voice recordings', Nissan's `medical data') exhibit high centrality of betweenness, highlighting their roles in facilitating connections among entities as well as amplifying the potential privacy risks if compromised.

\item \textbf{Third-party service providers' dependencies}: Audi, Ford, Nissan, Kia, Polestar, and Vauxhall demonstrate heavy reliance on third-party partners/service providers based on different network metrics, highlighting the complex accountability chains and potential attack surfaces.

\item \textbf{Cross-border data transmission}: Hyundai, Kia, and Lexus show cross-border data flows involving international entities. This raises privacy concerns when there are mismatches in data protection standards and legislation among different regions.

\item \textbf{Corporate-group centralised control}: Hyundai and Lexus display tighter intra-organisational control, with 60-75\% of their top-closeness nodes being affiliated entities. Different from the heavy reliance on third-party entities observed in other apps, this could potentially enable more standardised data governance.

\item \textbf{Transparency gaps}: Renault has `Unknown' nodes in different network metrics' top ten lists. With its generic approach of describing data flows, it reflects the transparency of its associated privacy policy. 
\end{itemize}

\subsubsection{Data Flow and Data Consumer Type Analysis}
\label{sec:pattern}

As shown in Table~\ref{tab:data_flow_stats}, among all automotive OEMs, Honda reported the highest number of data flows (as illustrated in Figure~\ref{fig:graph_comparison} (a)), whereas Renault reported only four (as illustrated in Figure~\ref{fig:graph_comparison} (b)). Assessing privacy practices solely based on the number of identified data flows is challenging because privacy policies vary significantly in writing style and level of detail. In addition, how well a privacy policy aligns with its actual data processing practices in reality is unknown. We acknowledge the limitations of using the privacy policy as a single source for conducting the analysis alone. However, if we assume that each privacy policy maintains a consistent style and structure throughout, analysing the normalised frequencies (i.e., relative proportions) of first-party, third-party, and incomplete data flows (see Table~\ref{tab:data_flows} in Section~\ref{subsubsec:post-processor_data_parser}) across different privacy policies can still provide meaningful insights. For each privacy policy, we calculated the fraction of each data flow type relative to the total number of data flows (values between 0 and 1). The results are depicted in Table~\ref{tab:data_flow_stats}.

\begin{table*}[!htb]
\centering
\caption{Data flow statistics across selected automotive OEMs}
\label{tab:data_flow_stats}
\resizebox{\linewidth}{!}{%
\begin{tabular}{lccccccccccc}
\toprule
& \textbf{Audi} & \textbf{Ford} & \textbf{Honda} & \textbf{Hyundai} & \textbf{Kia} & \textbf{Lexus} & \textbf{Nissan} & \textbf{Polestar} & \textbf{Renault} & \textbf{Vauxhall}\\ 
\midrule
\textbf{Number of data flows} & 56 & 109 & 348 & 65 & 63 & 45 & 141 & 80 & 4 & 142 \\
\midrule
\textbf{First-party data flow}\\
User-party to first-party & 0.17 & 0.19 & 0.38 & 0.20 & 0.37 & 0.22 & 0.33 & 0.025 & 0.50 & 0.31\\ 
First-party to first-party & 0.04 & 0.11 & 0.01 & 0.25 & 0.02 & 0.00 & 0.01 & 0.15 & 0.00 & 0.10\\ 
Third-party to first-party & 0.00 & 0.11 & 0.13 & 0.00 & 0.00 & 0.00 & 0.11 & 0.1 & 0.00 & 0.10\\ 
\textbf{Total} & \textbf{0.21} & \textbf{0.41} & \textbf{0.52} & \textbf{0.45} & \textbf{0.38} & \textbf{0.22} & \textbf{0.45} & \textbf{0.28} & \textbf{0.50} & \textbf{0.54}\\ 
\hline
\textbf{Third-party data flow}\\
User-party to third-party & 0.11 & 0.36 & 0.30 & 0.14 & 0.08 & 0.76 & 0.26 & 0.15 & 0.00 & 0.41\\ 
First-party to third-party & 0.36 & 0.08 & 0.05 & 0.35 & 0.54 & 0.00 & 0.24 & 0.125 & 0.00 & 0.01\\ 
Third-party to third-party & 0.20 & 0.06 & 0.01 & 0.00 & 0.00 & 0.00 & 0.01 & 0.00 & 0.00 & 0.03\\ 
\textbf{Total} & \textbf{0.66} & \textbf{0.51} & \textbf{0.36} & \textbf{0.49} & \textbf{0.62} & \textbf{0.76} & \textbf{0.51} & \textbf{0.28} & \textbf{0.00} & \textbf{0.45}\\ 
\hline
\textbf{Incomplete data flow} & \textbf{0.13} & \textbf{0.08} & \textbf{0.12} & \textbf{0.06} & \textbf{0.00} & \textbf{0.02} & \textbf{0.04} & \textbf{0.45} & \textbf{0.50} & \textbf{0.01}\\ 
\bottomrule
\end{tabular}
}
\begin{tablenotes}
\tiny
\item Values under ``Number of data flows'' are absolute counts; all other numerical values represent normalised frequencies of each data flow type relative to the total data flows within that OEM’s privacy policy.
\end{tablenotes}
\end{table*}

Across all data flows, each app follows a similar pattern, where the majority of first-party data flows is from user-party nodes to first-party nodes, whereas only a small fraction is from third-party nodes to first-party nodes. Overall, more than half of the data flows in Honda, Renault, and Vauxhall had a normalised frequency greater than 0.50 of their data flows classified as first-party. In contrast, Ford, Hyundai, Kia, and Nissan had values around 0.40, while Audi, Lexus, and Polestar showed even lower proportions of first-party data flows.

No clear pattern emerged for third-party data flows. Some apps (i.e., Ford, Honda, Lexus, Nissan, Vauxhall, Polestar) have a higher proportion of data flowing from the user-party to the third-party, while others (i.e., Audi, Hyundai, and Kia) have more flows from the first-party to the third-party. Overall, Audi, Ford, Kia, Lexus, and Nissan had more than half (normalised frequency <0.50) of their data flows classified as third-party, while the remaining apps fell between 0.20 and 0.50, except for Renault. Moreover, regarding the percentage of incomplete data flow, Polestar and Renault have a significantly higher proportion than the remaining apps, indicating potential unclear and ambiguity in their privacy policies.

To illustrate how such data can be used to gain more comprehensive insights and compare different privacy policies, we propose a generalised data flow risk score using max-normalisation of weights, aiming to provide indicative figures that can be used to compare different privacy policies in terms of potential risks and transparency. The score is computed by summing the weighted contributions of different data flow types and normalising the score to the range [0, 1]. As illustrated in Eq.~\eqref{eq:privacy_score}, the numerator of the equation represents the calculation of a raw privacy score for a given instance $i$, where $w_j$ is the weight for each data flow type and $X_{j,i}$ is the corresponding value for instance $i$. The denominator of the equation calculates the maximum possible score using the maximum observed values of each data flow type, where $\max(X_j)$ is the highest value observed for the $j$-th data flow type across all instances. Finally, the normalised data flow risk score was computed by dividing the raw score by the maximum possible score:
\begin{equation}
\label{eq:privacy_score}
S_{\text{norm}, i} = \frac{\sum_{j=1}^n w_j \cdot X_{j,i}}{\sum_{j=1}^n w_j \cdot \max(X_j)}
\end{equation}

It is worth noting that the selection of different weights below is based on our own experience and judgment. They are intended for illustrative purposes, and the subsequent analysis depends on these chosen weights. For first-party data flows, cases where data originate from third-party nodes and terminate at first-party nodes are assumed to pose more risks, as the data handling practices of third-party entities lie outside the control of both users and first parties. To reflect this, we apply weighted scores of 1 for user-party to first-party flows, 1.5 for first-party to first-party flows, and 2.25 for third-party to first-party flows. For third-party data flows, we apply weights based on the following assumption: 1 for user-party to third-party flows, 1.5 for first-party to third-party flows, and 2.25 for third-party to third-party flows. For the overall privacy scores, we assume that incomplete data flows pose a greater risk than both third-party and first-party data flows; hence, we assign a weight of 1 to overall scores of first-party data flows, 1.5 to third-party data flows, and 2.25 to incomplete data flows.

\begin{table*}[!htb]
\centering
\caption{Summary of data flow risk scores}
\label{tab:data_flow_privacy}
\resizebox{\linewidth}{!}{%
\begin{tabular}{lccccccccccc}
\toprule
& \textbf{APP1} & \textbf{APP2} & \textbf{APP3} & \textbf{APP4} & \textbf{APP5} & \textbf{APP6} & \textbf{APP7} & \textbf{APP8} & \textbf{APP9} & \textbf{APP10}\\
\midrule
Score for first-party data flows & 0.20 & 0.52 & 0.59 & 0.49 & 0.33 & 0.19 & 0.51 & 0.41 & 0.43 & 0.50\\
Score for third-party data flows & 0.54 & 0.31 & 0.24 & 0.33 & 0.44 & 0.38 & 0.32 & 0.17 & 0.00 & 0.25\\
Overall score & 0.54 & 0.49 & 0.47 & 0.48 & 0.47 & 0.51 & 0.47 & 0.61 & 0.58 & 0.49\\
\bottomrule
\end{tabular}
}
\end{table*}


It is important to acknowledge that privacy is a complex and multifaceted concept that cannot be fully captured using quantitative or qualitative rankings alone. Such measures can be misleading, or even harmful, if not considered alongside broader legal, organisational, and contextual factors. That is why we chose to anonymise apps' names in this part of the analysis, as it is mainly intended to illustrate methodological insights rather than to systematically and comprehensively rank privacy risks.

As reported in Table~\ref{tab:data_flow_privacy}, a higher score represents higher data flow-related risks. APP2, APP3, APP4, APP7, APP9, and APP10 have relatively higher risk scores for first-party data flow (i.e., between 0.43 and 0.59) than their scores for third-party data flows. In contrast, APP1, APP5, APP6, and APP8 have higher risk scores for third-party data flows, with APP1 having the highest risk (i.e., 0.54), followed closely by APP5 and APP6 with scores of 0.44 and 0.38, respectively. It is worth noting that APP9 has a score of 0.00, indicating no third-party data flow risks, which is due to the absence of such data flows. This is shown in Figure~\ref{fig:graph_comparison}, suggesting that the privacy policy is overly broad and lacks the granularity needed to better inform users about their general data handling practices. 

In addition, with the contribution of incomplete data flows to the overall risk scores, APP1, APP6, APP8, and APP9 had the highest scores, approximately between 0.51 and 0.61. This aligns with their higher percentage of incomplete data flows, indicating ambiguity and a lack of transparency in privacy policies. This is also in line with the observations obtained from the graph analysis presented in Section~\ref{subsubsec:case_study_results_network_analysis}, where they were singled out for potential privacy concerns related to their dependency on third-party entities and privacy policy transparency issues. The overall risk scores for the remaining apps fell just below 0.50, indicating that these apps have relatively better or clearer data handling practices stated in their privacy policies from the data flow perspective.

\subsubsection{Data Category and Data Processing Purpose Analysis}

One of the main tasks for the LLM-based processor is to further classify data types into data categories based on the definitions set in knowledge typology $KT_\text{data}$ to allow more in-depth analysis. As depicted in Table~\ref{tab:data_category_distribution}, Honda stands out with the highest number of data types across almost all data categories. Honda collects more data types, particularly in categories such as `Online identifiers' (25), `User online activities' (18), `Location' (13), and `Finance' (17), indicating its broad data processing related to online user tracking and financial transactions. However, Audi, Kia, Lexus, Polestar, Renault, and Vauxhall collected significantly less data across most data-type categories. In addition, the `Other' data type category contains data types that cannot be categorised using the other categories listed in the table. By manually inspecting this data category, it mostly contains vehicle-related data (e.g., `speed', `vehicle data', `braking' and `DTC (Diagnostic Trouble Code) history') and other unspecified data (e.g., `reason for your Roadside Assistance call', `timestamp of the generated security event').

\begin{table}[!htb]
\centering
\caption{Data category distribution across selected automotive companies}
\label{tab:data_category_distribution}
\resizebox{\linewidth}{!}{%
\begin{tabular}{lcccccccccc}
\toprule
\textbf{Data Category} & \textbf{Audi} & \textbf{Ford} & \textbf{Honda} & \textbf{Hyundai} & \textbf{Kia} & \textbf{Lexus} & \textbf{Nissan} & \textbf{Polestar} & \textbf{Renault} & \textbf{Vauxhall}\\
\midrule
Generic Personal Info & 1 & 5 & 7 & 4 & 3 & 3 & 9 & 5 & 1 & 2\\
Personal ID Identifier & 1 & 3 & 4 & 0 & 2 & 1 & 2 & 1 & 0 & 0\\
Online Identifier & 2 & 3 & 25 & 0 & 3 & 1 & 4 & 7 & 2 & 6\\
User Online Activities & 9 & 5 & 18 & 0 & 2 & 0 & 12 & 7 & 0 & 6\\
User Profile & 0 & 1 & 1 & 0 & 0 & 0 & 1 & 0 & 0 & 1\\
Contact & 1 & 7 & 8 & 2 & 2 & 4 & 4 & 5 & 0 & 5\\
Demographic & 0 & 3 & 13 & 2 & 0 & 1 & 4 & 5 & 0 & 2\\
Biometric Information & 0 & 0 & 1 & 0 & 2 & 1 & 1 & 0 & 1 & 0\\
Location & 1 & 15 & 13 & 0 & 10 & 1 & 1 & 8 & 0 & 1\\
Finance & 0 & 2 & 17 & 0 & 0 & 0 & 12 & 5 & 0 & 0\\
Health & 0 & 1 & 0 & 0 & 0 & 0 & 1 & 3 & 0 & 0\\
Criminal Records & 0 & 2 & 0 & 0 & 0 & 0 & 0 & 0 & 0 & 0\\
Device Information & 2 & 6 & 11 & 1 & 2 & 0 & 1 & 9 & 0 & 5\\
Other & 7 & 9 & 60 & 5 & 6 & 3 & 17 & 10 & 0 & 7\\
\bottomrule
\end{tabular}%
}
\end{table}

Certain data categories, such as `Profile', `Biometric Information', `Finance', `Health', and `Criminal Records', exhibit clear variations between apps. `Biometric Information' is collected by multiple apps, including Honda (i.e., voice command), Kia (i.e., voice recording, voice samples), Lexus (i.e., sound or images files), Nissan (i.e., voice recordings from calls between you and the dealer), and Renault (i.e., information that makes it possible to identify you). A broad range of `Finance' data is collected by Honda (e.g., tax code, payment card data, PAN, cardholder name, transaction information) and Nissan (e.g., purchasing history, payment method, discount granted, order history, order number). `Health' data processing for Nissan is medical personal data.  Meanwhile, Ford is the only company that explicitly claims a collection of `Criminal Records', including crime reference numbers and vehicle theft information. Regarding `User Profile' data, Ford collects profile pictures, Honda records information about the primary user, Nissan collects usernames and profiling details, and Vauxhall claims the processing of general profile data. Overall, the differences in data processing across apps presented in Table~\ref{tab:data_category_distribution} indicate varying levels of specificity and transparency in privacy policies, with some apps explicitly detailing specific data types, while others remain more general or omit such disclosures.

\begin{figure}[!htb]
\centering
\begin{adjustbox}{width=\linewidth}
\begin{tikzpicture}
\begin{axis}[
    ybar stacked,
    bar width=12pt,
    ymin=0,
    ymax=1,
    ylabel={Proportion},
    enlarge x limits=0.1,
    enlarge y limits=0.1,
    symbolic x coords={
        Contacts,
        Demographics,
        Device information,
        Generic personal information,
        Location,
        Online identifier,
        User online activities,
        Personal identity identifier
    },
    xtick=data,
    y tick label style={font=\scriptsize},
    x tick label style={font=\scriptsize, rotate=45, anchor=east},
    legend columns=2,
    legend style={
        font=\scriptsize,
        at={(1.1, 0.5)},
        anchor=west,
        draw=none,
        /tikz/every even column/.append style={column sep=0.5cm}
    },
    area legend
]

\addplot+[fill=blue!20] coordinates {
    (Contacts,0.077669903) (Demographics,0.016666667) (Device information,0.06741573) (Generic personal information,0.047029703)
    (Location,0.101123596) (Online identifier,0.070707071) (User online activities,0.025) (Personal identity identifier,0)
};

\addplot+[fill=red!20] coordinates {
    (Contacts,0.029126214) (Demographics,0.016666667) (Device information,0.02247191) (Generic personal information, 0)
    (Location,0.04494382) (Online identifier, 0) (User online activities,0.041666667) (Personal identity identifier,0)
};

\addplot+[fill=green!30] coordinates {
    (Contacts,0.009708738) (Demographics,0.033333333) (Device information,0.168539326) (Generic personal information,0.024752475)
    (Location,0.112359551) (Online identifier,0.121212121) (User online activities,0.341666667) (Personal identity identifier,0)
};

\addplot+[fill=orange!30] coordinates {
    (Contacts,0.32038835) (Demographics,0.058333333) (Device information,0.14607416) (Generic personal information,0.165841584)
    (Location,0.123595506) (Online identifier,0.121212121) (User online activities,0.141666667) (Personal identity identifier,0.222222222)
};

\addplot+[fill=purple!30] coordinates {
    (Contacts,0.126213592) (Demographics,0.303833333) (Device information,0.179752281) (Generic personal information,0.188118812)
    (Location,0.191011236) (Online identifier,0.202020202) (User online activities,0.041666667) (Personal identity identifier,0.185185185)
};

\addplot+[fill=yellow!40] coordinates {
    (Contacts,0.116504854) (Demographics,0.016666667) (Device information,0.04494382) (Generic personal information,0.014851485)
    (Location,0.011235955) (Online identifier,0.03030303) (User online activities,0) (Personal identity identifier,0.111111111)
};

\addplot+[fill=cyan!30] coordinates {
    (Contacts,0.087378641) (Demographics,0) (Device information,0.02247191) (Generic personal information,0.183168317)
    (Location,0.08988764) (Online identifier,0.02020202) (User online activities,0.05) (Personal identity identifier,0.185185185)
};

\addplot+[fill=brown!30] coordinates {
    (Contacts,0.174757282) (Demographics,0.483333333) (Device information,0.303370787) (Generic personal information,0.188118812)
    (Location,0.269662921) (Online identifier,0.414141414) (User online activities,0.241666667) (Personal identity identifier,0.296296296)
};

\addplot+[fill=gray!40] coordinates {
    (Contacts,0) (Demographics,0.033333333) (Device information,0.033707865) (Generic personal information,0.131188119)
    (Location,0.033707865) (Online identifier,0) (User online activities,0.116666667) (Personal identity identifier,0)
};

\addplot+[fill=yellow!70] coordinates {
    (Contacts,0.058252427) (Demographics,0.033333333) (Device information,0.011235955) (Generic personal information,0.056930693)
    (Location,0.02247191) (Online identifier,0.02020202) (User online activities,0) (Personal identity identifier,0)
};

\legend{
    Additional Service or Feature,
    Advertising,
    Analytics or Research,
    Basic Service or Feature,
    Legal Requirement,
    Marketing,
    Merger/Acquisition,
    Operational Integrity and Security,
    Personal Data,
    Social Media Integration,
    Other (Upsell)
}
\end{axis}
\end{tikzpicture}
\end{adjustbox}
\caption{Distribution of data processing purposes across different data categories.}
\label{fig:purpose_distribution}
\end{figure}
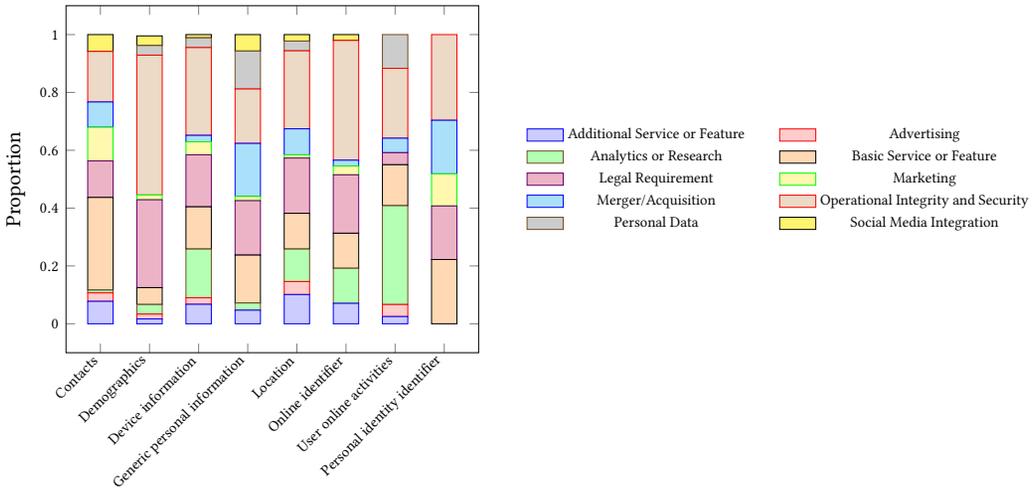

Furthermore, we were interested in how different data categories are associated with different data processing purposes. To achieve this, we hand-picked data categories that were claimed to be collected and shared by two-thirds of the apps and examined the distribution of data processing purposes across different data categories collectively. Overall, as shown in Figure~\ref{fig:purpose_distribution}, almost all data categories are linked to multiple purposes of data processing to some extent. This highlights the multifunctional role that personal data plays in the automotive context, serving both operational needs and broader objectives.

Among all data categories, `Operational Integrity and Security' is the most frequently associated data processing purpose across most data categories. This is partly due to its general and broad definition, making it applicable to a wide range of data flows. In addition, data processing purpose `Legal Requirement' are consistently prominent across most of data categories. Combined with `Operational Integrity and Security', this might mean that legal and compliance-based justifications underpin much of the identified data categories.

Moreover, multiple data categories such as `Contacts', `Generic Personal Information', `Location', `Online Identifier', `User Online Activities' and `Device Information' stand out with a diverse spread, where they are frequently associated with a wide range of data processing purposes, indicating their multiple roles in the data processing. For instance, `Contacts', its association with multiple data processing purposes suggests its dual role in both functional (i.e., basic service, communication) and business-oriented uses (i.e., marketing and promotion). `Location', `Device Information', `Online Identifier', and `User Online Activities' show the strong link to the purpose `Analytics or Research', reflecting their additional role in possible behavioural profiling and service optimisation. 

The processing of `Personal Identity Identifier' category is commonly linked with purposes including `Basic Service or Feature', `Legal Requirement', `Marketing', and `Merger/Acquisition', in addition to `Operational Integrity and Security', indicating the broad scale of sensitive data processing, and raising potential privacy concerns about oversharing personal sensitive data.

In general, mapping data processing purposes to data categories provides insight into the collective behaviour of data-handling practices within the automotive industry. This further confirms and raises potential concerns about necessary data minimisation, excessive data retention, and potential secondary use. Moreover, it would be interesting to take privacy policies from other domains and conduct cross-sector comparison studies; however, this is out of the scope of this paper and could be an interesting topic as part of our future work.

\subsubsection{Cross-Check with Additional Data Sources}

Although privacy policies are subject to legal requirements to disclose data-handling practices, the extent to which they provide accurate information is unclear. Hence, we need other sources of information to cross-check the results. To this end, we visited the Google Play (i.e., Android) apps' data safety sections\footnote{\url{https://support.google.com/googleplay/answer/11416267}} and the Apple iOS apps' privacy labels\footnote{\url{https://developer.apple.com/app-store/app-privacy-details/}}, where both require app developers to disclose information about mobile apps' data collection and data sharing practices.

As shown in Tables~\ref{tab:google_data} and \ref{tab:apple_data} in Section~\ref{sec:appendix}, across all selected mobile apps, disclosures via both the Google Play apps' data safety sections and Apple iOS apps' privacy labels contradict the data practices stated in their associated privacy policies. For instance, the NissanConnect Services app collects financial data as stated in its privacy policy, but it fails to declare this in its Google Play app's data safety section and Apple iOS apps' privacy labels. Moreover, even for the same mobile app, there are noticeable inconsistencies in data collection and sharing disclosures between Google Play's app safety section and Apple iOS apps' privacy labels. For instance, Audi's app does not collect any data according to its Apple iOS app's privacy labels, whereas several types of personal information are collected based on its Google Play app's safety description. This may suggest that 1) mobile developers may be negligent in accurately reporting their data-handling practices; 2) there is a lack of a clear understanding of the regulatory landscape; and 3) there is a significant discrepancy in data security and privacy between Google Play apps and Apple iOS apps. These observations highlight the difficulty in verifying the accuracy of privacy policies, while also revealing the unreliability of Google Play apps' data safety sections and Apple iOS apps' privacy labels. The significant discrepancies between different data sources raise concerns about the extent to which consumers and, perhaps, even developers can truly understand the full scope and granularity of data-handling practices. However, systematically addressing these is beyond the scope of this study, but it is worth investigating in future studies.

\section{Limitations and Future Work}
\label{sec:limitations}

One limitation of this work is the verification process. The lack of an existing ground-truth dataset specifically designed for extracting data flows from text prevents us from conducting a large-scale evaluation. This study relied only on manual verification involving human verifiers to validate the extracted information. While this helps confirm and ensure the accuracy of our methods, it inherently limits our ability to benchmark the results with other methods. To address this limitation, \emph{we plan to utilise LLMs teaming with human users to co-curate dedicated ground truth datasets as part of our future work}. In addition, we acknowledge that relying on three co-authors as evaluators can introduce potential biases, which could potentially affect the generalisability and robustness of the validation results. To address this limitation, \emph{we are planning to conduct a more independent validation study with recruited human participants as part of our future work.}

It is also important to acknowledge that LLMs could be employed not only within the analyser but also for the pre-processor and data flow post-processor of the proposed framework. For instance, LLMs could be utilised to assist in refining input format, segmenting text, or enhancing insights discovery. However, in this study we restricted the use of LLMs to the analyser component, as the analyser represents the most complex and demanding stage of the pipeline, where the benefits of integrating LLMs are expected to be most promising. While this design choice allowed us to focus on demonstrating the feasibility and effectiveness of \SysName, it also constitutes a limitation. Future research could explore and evaluate the use of LLMs to support to other components of the framework.

Moreover, we consider that the construction of the knowledge bases used in the RAG implementation is another limitation of this work, since the accuracy of these knowledge bases can affect the results generated by the LLMs. Our findings demonstrate that vague or unclear definitions can lead to ambiguous and incorrect outputs from LLMs. For instance, concepts such as active and passive/automatic data processing are inherently difficult to define with great clarity. Even for human readers, the interpretations of the same text by different people would be different. To this end, we would recommend \emph{that knowledge bases should be co-created with domain experts to enhance clarity and improve the reliability of LLMs' outputs while also helping to reduce the risk of hallucinations.}

As introduced in Section~\ref{subsubsec:pre-processor_Module_TS}, LLMs process text segments sequentially to complete a set of tasks. This approach may overlook cross-paragraph (i.e., cross-segment) information, potentially leading to unreliable or incomplete outputs from LLMs. However, we found that for a granular task such as extracting data flows, feeding the entire privacy policy as the input would lead to 1) the loss of granularity compared to our current approach of sequentially feeding smaller text segments and 2) incorrect association of unrelated data senders/data receivers with data types. Nevertheless, we would like to emphasise that we also made some efforts to mitigate this limitation by using paragraph-based text segmentation, adding headings to bullet point-based segments, and adding a table header row to each table row for additional contextual information. In a nutshell, we would \emph{recommend that further research is needed to 1) determine the optimal text segmentation approach with considerations of cross-paragraph contexts to improve LLMs' performance on such fine-grained tasks and 2) explore other approaches beyond RAG such as contextual-retrieval\footnote{\url{https://www.anthropic.com/news/contextual-retrieval}} and knowledge augmented generation (KAG)~\cite{Liang-L2024}}.

Furthermore, our current network graphs of data flows reveal strong connections between corporate groups and third parties. It would be useful to allow the propose framework to integrate with existing tools (e.g., NetVizCorpy~\cite{Baruwa-Z2025}, a tool that reconstructs business‑to‑business relationship graphs using Wikidata) to enrich and expand the scope of our network analysis by providing a more comprehensive view of inter‑organisational data flows extracted from other data sources, as well as helping validating the outputs generated by \SysName.

Last but not least, while this work focuses on analysing privacy policies for the automotive industry, the aim is to demonstrate its capability and effectiveness of conducting complicated tasks using LLMs with RAG. We would also like to \emph{highlight that \SysName\ is not only capable of analysing privacy policies but also generalisable and adaptable for examining any text-based documents for other domains}. Additionally, many of the parts, such as the text segmentation tool, domain knowledge bases, and LLM-based agents in the proposed framework, \emph{can be updated and customised for different tasks to provide better flexibility and generalisability.} In our future work, we will explore applications of \SysName\ to different types of documents and different tasks.

\section{Conclusion}
\label{sec:conclusion}

This paper reports our work on developing and evaluating an end-to-end framework, \SysName, for automating privacy policy analysis using LLMs and RAG. \SysName\ consists of a pre-processor, an LLM-based processor, and a data flow post-processor. It can 1) preprocess privacy policy texts by converting and segmenting HTML-based text and constructing local knowledge bases; 2) utilise several LLM agents to analyse privacy policy text segments and extract comprehensive data flows; and 3) post-process LLMs' outputs and utilise data flow graphs to discover data protection and privacy-related insights. To demonstrate the usefulness and effectiveness of the proposed framework, we conducted a case study involving the analysis of privacy policies from ten selected connected vehicle mobile apps. The results show that \SysName\ can effectively and accurately understand privacy policies and extract key information to generate comprehensive data flows, as well as reveal privacy insights that may require considerable time for consumers to read and comprehend.

\bibliographystyle{ACM-Reference-Format}
\bibliography{main}
\newpage

\section*{Appendix}
\label{sec:appendix}

\begin{figure}[!htb]
\centering
\begin{tcolorbox}[colback=white, colframe=black, colbacktitle=black, title={Prompt}]
\textcolor{blue}{role: system} 

\textcolor{blue}{content:} 
\begin{enumerate}[leftmargin=0.8cm,labelsep=-0.1cm,align=left,label={[\arabic*]}]
\item You are an expert to analyse the \CAP{TEXT SEGMENT} to extract data flows.

\item If \CAP{TEXT SEGMENT} starts with \texttt{\_table\_}: Treat \texttt{|} as separators; Treat first line as the table heading; Treat second line as the table content.

\item Read and understand the \CAP{TEXT SEGMENT} and then strictly follow the below rules to produce your responses:
    \begin{enumerate} [leftmargin=0.7cm,labelsep=-0.2cm,align=left]
        \item If the \CAP{TEXT SEGMENT} at least talk about one party collects data or personal information from another party, or a party shares data or personal information to another party, OUTPUT the extracted data flows in multiple JSON objects.
        \item The JSON objects must use the format: 
        \item[] \begin{lstlisting}[language=json]
        {"Output":[{
            "data_sender": "",
            "data_type": [],
            "data_receiver": []
        }]}
        \end{lstlisting}
        \item Respond only with valid JSON. 
        \item Each data flow represents one party (i.e.,data\_receiver) collects personal data (i.e., data\_type) from another party (i.e., data\_sender), or a party (i.e., data\_sender) shares personal data (i.e., data\_type) to another party (i.e., data\_receiver).
    
        \item For data\_types: extract all atomic personal data (i.e., data\_type) following these rules: (1) when dealing with sentences that have combined data\_types, split them into individual data\_types for a clearer representation; (2) each data\_type MUST appear in \CAP{TEXT SEGMENT}, and do not change the cases; (3) DO NOT INCLUDE any other text in the answer such as input or query text or your deduction or your explanation; (4) remove Pronouns in the identified strings; (5) DO NOT INCLUDE specific addresses, postcodes, email addresses, companies, organisations, or geographical information; (6) if you can not identify a data\_type, leave it empty.
        
        \item For data\_sender/data\_receiver: (1) when dealing with sentences that have combined data\_receivers or data\_senders, split them into individual data\_receiver or data\_sender for a clearer representation; (2) each data\_sender or data\_receiver string MUST appear in \CAP{TEXT SEGMENT}, do not change the cases; (3) if no data\_sender is explicitly stated in the \CAP{TEXT SEGMENT}, leave data\_sender empty; (4) if no data\_receiver is explicitly stated in the \CAP{TEXT SEGMENT}, leave data\_receiver empty. 
        \item OUTPUT only ``None'' for other unrelated scenarios
    \end{enumerate}
\end{enumerate}

\textcolor{blue}{role: user}

\textcolor{blue}{content:} 
\begin{itemize} [leftmargin=0.8cm,labelsep=-0.1cm,align=left]
\item[] \CAP{TEXT SEGMENT}: One Privacy Policy Text Segment  
\end{itemize}
\end{tcolorbox}
\caption{Prompt template for extracting data flows following the Groq Chat Completions API format}
\label{fig:prompt_dataflow}
\end{figure}

\begin{figure}[!htb]
\centering
\begin{tcolorbox}[colback=white, colframe=black, title=Prompt]
\textcolor{blue}{role: system} 
\textcolor{blue}{content:} 
\begin{enumerate}[leftmargin=0.8cm,labelsep=-0.1cm,align=left,label={[\arabic*]}]
\item You are an expert in categorising the \CAP{INPUT DATA TYPE} provided by the user to extract key information about data collection/sharing.

\item Use the \CAP{TEXT SEGMENT} given by the user, along with the knowledge and understanding of data category and description provided in the \CAP{CONTEXT} list to perform the categorisation.

\item Please strictly follow the below rules when answering questions:
    \begin{enumerate} [leftmargin=0.7cm,labelsep=-0.2cm,align=left]
        \item Output must follow the JSON format shown below.  
        \begin{lstlisting}[language=json]
        {"Output": [{
            "DataCategory": "data_category",
            "DataType": "data_type",
            "InputText": "input_text",
        }]}
        \end{lstlisting}
        \item Respond only with valid JSON. 
        \item Do not include any other text (e.g., input or query text).
        \item DataCategory is the identified data category defined in \CAP{CONTEXT}. Do not create new categories.  
        \item DataType is the \CAP{INPUT DATA TYPE}.  
        \item InputText is the \CAP{TEXT SEGMENT}.  
        \item OUTPUT only "None", if no category can be found. 
    \end{enumerate}
\end{enumerate}

\textcolor{blue}{role: user} 
\textcolor{blue}{content:} 
\begin{enumerate}[leftmargin=0.8cm,labelsep=-0.1cm,align=left,label={[\arabic*]}]
\item \CAP{INPUT DATA TYPE}: Example Data Type

\item \CAP{TEXT SEGMENT}: Privacy Policy Text Segment

\item \CAP{CONTEXT}:
    \begin{itemize}[leftmargin=0.8cm,labelsep=-0.1cm,align=left]
        \item[] Data category: Data category 1
        \item[] Data description: Description of data category 1
        \item[] Example data types: data1, data2, data3, ...
    \end{itemize}
    
\item \CAP{CONTEXT}:
    \begin{itemize}[leftmargin=0.8cm,labelsep=-0.1cm,align=left]
        \item[] Data category: Data category 2
        \item[] Data description: Description of data category 2
        \item[] Example data types: data1, data2, data3, ...
    \end{itemize}
\end{enumerate}
\end{tcolorbox}
\caption{Prompt template for identifying data categories following the Groq Chat Completions API format}
\label{fig:prompt_data_cateogry}
\end{figure}

\begin{table*}[!htb]
\centering
\caption{Data collection and sharing practices stated in Google app safety section}
\label{tab:google_data}
\begin{threeparttable}
\resizebox{\linewidth}{!}{%
\begin{tabular}{|l|cc|cc|cc|cc|cc|cc|cc|cc|cc|cc|}
\hline
\textbf{Data Category} & \multicolumn{2}{c|}{\textbf{Audi}} & \multicolumn{2}{c|}{\textbf{Ford}} & \multicolumn{2}{c|}{\textbf{Honda}} & \multicolumn{2}{c|}{\textbf{Hyundai}} & \multicolumn{2}{c|}{\textbf{Kia}} & \multicolumn{2}{c|}{\textbf{Lexus}} & \multicolumn{2}{c|}{\textbf{Nissan}} & \multicolumn{2}{c|}{\textbf{Polestar}} & \multicolumn{2}{c|}{\textbf{Renault}} & \multicolumn{2}{c|}{\textbf{Vauxhall}} \\
\hline
& S & C & S & C & S & C & S & C & S & C & S & C & S & C & S & C & S & C & S & C\\ 
\hline
\textbf{Location} &  & & & & & & & & & & & & & & & & & & &\\
Approximate location & & & & & & & & & & & & & & & & & Y & & Y &\\
Precise location & & Y & & & & & & & & & & & & & & & Y & & Y &\\
\hline
\textbf{Personal Info} & & & & & & & & & & & & & & & & & & & &\\
Name & & Y & & Y & Y & Y & & Y & & Y & Y & Y & & & & Y & Y & Y & Y & Y\\
Email address & & Y & & Y & Y & Y & & Y & & Y & Y & Y & & Y & Y & Y & Y & Y & Y & Y\\
User IDs & Y & Y & & Y & Y & Y & & & & & & Y & & & Y & Y & Y & Y & Y &\\
Address & & Y & & Y & Y & Y & & & & & Y & Y & & & & & Y & Y & & Y\\
Phone number & & Y & & Y & & & & Y & & Y & Y & Y & & & Y & Y & Y & Y & Y & Y\\
Race and ethnicity & & & & & & & & & & & & & & & & & & & &\\
Political or religious beliefs & & & & & & & & & & & & & & & & & & & &\\
Sexual orientation & & & & & & & & & & & & & & & & & & & &\\
Other info & & & & & & Y & & & & & & & & & & & Y & Y & Y & Y\\
\hline
\textbf{Financial Info} & & & & & & & & & & & & & & & & & & & &\\
User payment info & & & & & & & & & Y & & & & & & & & & & &\\
Purchase history & & & & & & & & & & & & & & & & & & & &\\
Credit score & & & & & & & & & & & & & & & & & & & &\\
Other financial info & & & & & & & & & & & & & & & & & & & &\\
\hline
\textbf{Health \& Fitness} & & & & & & & & & & & & & & & & & & & &\\
Health info & & & & & & & & & & & & & & & & & & & &\\
Fitness info & & & & & & & & & & & & & & & & & & & &\\
\hline
\textbf{Messages} & & & & & & & & & & & & & & & & & & & &\\
Emails & & & & & & & & & & & & & & & & & & & &\\
SMS or MMS & & & & & & & & & & & & & & & & & & & &\\
Other in-app messages & & & & & & & & & & & & & & & & & & & &\\
\hline
\textbf{Photos and Videos} & & & & & & & & & & & & & & & & & & & &\\
Photos & & & & & & & & & & Y & Y & & & & & Y & & & &\\
Videos & & & & & & & & & & & & & & & & & & & &\\
\hline
\textbf{Audio files} & & & & & & & & & & & & & & & & & & & &\\
Voice or sound recordings & & & & & & & & & & & & & & & & & & & &\\
Music files & & & & & & & & & & & & & & & & & & & &\\
Other audio files & & & & & & & & & & & & & & & & & & & &\\
\hline
\textbf{Files and docs} & & & & & & & & & & & & & & & & & & & &\\
Files and docs & & & & & & & & Y & & Y & & & & & & & & & Y &\\
\hline
\textbf{Calendar} & & & & & & & & & & & & & & & & & & & &\\
Calendar events & & Y & & & & & & & & & & & & & & & & & &\\
\hline
\textbf{Contacts} & & & & & & & & & & & & & & & & & & & &\\
Contacts & & & & & & & & Y & & Y & & & & & & & & & &\\
\hline
\textbf{App activity} & & & & & & & & & & & & & & & & & & & &\\
App interactions & & Y & & Y & Y & Y & & Y & & & Y & & & & & & Y & & Y & Y\\
In-app search history & & & & & & & & & & & & & & & & & & & &\\
Installed apps & & & & & & & & & & & & & & & & & & & &\\
Other user-generated content & & & & & Y & Y & & & & & Y & & & & & Y & & & &\\
Other actions & & & & & & & & & & & & & & & & & & & Y & Y\\
\hline
\textbf{Web browsing} & & & & & & & & & & & & & & & & & & & &\\
Web browsing history & & & & & & & & & & & & & & & & & & & &\\
\hline
\textbf{App info and performance} & & & & & & & & & & & & & & & & & & & &\\
Crash logs & & & & Y & Y & Y & & Y & & Y & Y & & & & & Y & Y & & Y &\\
Diagnostics & & & & Y & Y & Y & & Y & & Y & & Y & & & & & Y & & Y &\\
Other app performance data & & & & & & & & & & & & Y & & & & & Y & & &\\
\hline
\textbf{Device or other IDs} & & & & & & & & & & & & & & & & & & & &\\
Device or other IDs & & Y & & Y & Y & Y & & Y & & Y & Y & Y& & & & Y & Y & Y & Y & Y\\
\hline
\end{tabular}
}
\begin{tablenotes}
\footnotesize
\item S: Data shared
\item C: Data collected
\end{tablenotes}
\end{threeparttable}
\end{table*}

\begin{table*}[!htb]
\centering
\caption{Data collection and sharing practices represented using Apple privacy labels}
\label{tab:apple_data}
\begin{threeparttable}
\resizebox{\linewidth}{!}{%
\begin{tabular}{|l|ccc|ccc|ccc|ccc|ccc|ccc|ccc|ccc|ccc|ccc|}
\hline
\textbf{Data Category} & \multicolumn{3}{c|}{\textbf{Audi}} & \multicolumn{3}{c|}{\textbf{Ford}} & \multicolumn{3}{c|}{\textbf{Honda}} & \multicolumn{3}{c|}{\textbf{Hyundai}} & \multicolumn{3}{c|}{\textbf{Kia}} & \multicolumn{3}{c|}{\textbf{Lexus}} & \multicolumn{3}{c|}{\textbf{Nissan}} & \multicolumn{3}{c|}{\textbf{Polestar}} & \multicolumn{3}{c|}{\textbf{Renault}} & \multicolumn{3}{c|}{\textbf{Vauxhall}}\\
\hline
& T & L & N & T & L & N & T & L & N & T & L & N & T & L & N & T & L & N & T & L & N & T & L & N & T & L & N & T & L & N\\
\hline
\textbf{Location} & & & & & & & & & & & & & & & & & & & & & & & & & & & & & &\\
Precise location & & & & & & & & & & & & & & Y & & & & Y & & & Y & & Y & & & & Y & & & Y\\
Coarse location & & & & & & & & & & & & Y & & & & & & & & & Y & & Y & & & & Y & & &\\
\hline
\textbf{Contact info} & & & & & & & & & & & & & & & & & & & & & & & & & & & & & &\\
Name & & & & & Y & & & Y & & & Y & & & Y & & & & & & & Y & & Y & & & Y & & & Y &\\
Email address & & & & & Y & & & Y & & & Y & & & Y & & & Y & & & Y & & & Y & & & Y & & & Y &\\
Phone number & & & & & Y & & & Y & & & Y & & & Y & & & Y & & & Y & & & Y & & & & & & Y &\\
Physical address & & & & & Y & & & Y & & & & & & & & & Y & & & & Y & & Y & & & & & & Y &\\
Other user contact info & & & & & & & & & & & & & & & & & Y & & & Y & & & & & & & & & &\\
\hline
\textbf{Health and fitness} & & & & & & & & & & & & & & & & & & & & & & & & & & & & & &\\
Health & & & & & & & & & & & & & & & & & & & & & & & & & & & & & &\\
Fitness & & & & & & & & & & & & & & & & & & & & & & & & & & & & & &\\
\hline
\textbf{Financial info} & & & & & & & & & & & & & & & & & & & & & & & & & & & & & &\\
Payment info & & & & & & & & & & & & & & & & & & & & & & & Y & & & & & & &\\
Credit info & & & & & & & & & & & & & & & & & & & & & & & Y & & & & & & &\\
Other financial info & & & & & & & & & & & & & & & & & & & & & & & & & & & & & &\\
\hline
\textbf{Sensitive info} & & & & & & & & & & & & & & & & & & & & & & & & & & & & & &\\
\hline
\textbf{Contacts} & & & & & & & & & & & & & & & & & & & & & & & & & & & & & &\\
\hline
\textbf{User content} & & & & & & & & & & & & & & & & & & & & & & & & & & & & & &\\
Emails or text messages & & & & & & & & & & & & & & & & & & & & & & & & & & & & & &\\
Photos or videos & & & & & & & & & & & & & & & & & & Y & & & & & & & & & & & & Y\\
Audio data & & & & & & & & & & & & & & & & & & & & & & & & & & & & & &\\
Gameplay content & & & & & & & & & & & & & & & & & & & & & & & & & & & & & &\\
Customer support & & & & & Y & & & & & & & & & & & & & & & & & & Y & & & & & Y & Y &\\
Other user content & & & & & & & & & & & & & & & & & & & & & & & Y & & & & & & &\\
\hline
\textbf{Browsing history} & & & & & & & & & & & & & & & & & & & & & & & & & & & & & &\\
\hline
\textbf{Search history} & & & & & & & & & & & & & & & Y & & & & & & & & & & & & & & & Y\\
\hline
\textbf{Identifiers} & & & & & & & & & & & & & & & & & & & & & & & & & & & & & &\\
User ID & & & & & Y & & & Y & & & & & & Y & & & & Y  & & & & & Y & & Y & Y & & & Y &\\
Device ID & & & & & Y & & & Y & & & & & & Y & & & & & & & Y & Y & Y & & & & & & Y &\\
\hline
\textbf{Purchase history} & & & & & & & & & & & & & & & & & & & & & & & Y & & & & & & &\\
\hline
\textbf{Usage data} & & & & & & & & & & & & & & & & & & & & & & & & & & & & & &\\
Product interaction & & & & & Y & & & Y & & & & & & & & & & Y & Y & & Y & & Y & & & & Y & & & Y\\
Advertising data & & & & & & & & & & & & & & & & & & & & & & & & Y & & & & & & Y\\
Other usage data & & & & & & & & & & & & & & & & & & & & & Y & & & & & & & & &\\
\hline
\textbf{Diagnostics} & & & & & & & & & & & & & & & & & & & & & & & & & & & & & &\\
Crash data & & & & & Y & & & Y & & & & & & & & & & Y & & & Y & & & Y & & & Y & & & Y\\
Performance data & & & & & Y & & & Y & & & & & & & & & & Y & & & & & & Y & & & & & & Y\\
Other diagnostic data & & & & & & & & Y & & & & & & & & & & & & & & & & & & & & & &\\
\hline
\textbf{Surroundings} & & & & & & & & & & & & & & & & & & & & & & & & & & & & & &\\
Environment scanning & & & & & & & & & & & & & & & & & & & & & & & & & & & & & &\\
\hline
\textbf{Body} & & & & & & & & & & & & & & & & & & & & & & & & & & & & & &\\
Hands & & & & & & & & & & & & & & & & & & & & & & & & & & & & & &\\
Head & & & & & & & & & & & & & & & & & & & & & & & & & & & & & &\\
\hline
\textbf{Other data types} & & & & & & & & Y & & & & & & Y & & & & & & & & & & & & & & & &\\
\hline
\end{tabular}
}
\begin{tablenotes}
\footnotesize
\item T: ``Data used to track you''
\item L: ``Data linked to you''
\item N: ``Data not linked to you''
\end{tablenotes}
\end{threeparttable}
\end{table*}

\end{document}